\pdfoutput=1

\documentclass[11pt]{article}

\usepackage[preprint]{acl}

\usepackage{times}
\usepackage{latexsym}

\usepackage[T1]{fontenc}
\usepackage[utf8]{inputenc}
\usepackage{microtype}
\usepackage{hyperref}       
\usepackage{url}            
\usepackage{booktabs}       
\usepackage{amsfonts}       
\usepackage{nicefrac}       
\usepackage{microtype}      
\usepackage{xcolor}         
\usepackage{amsmath}
\usepackage{amssymb}
\usepackage{mathtools}
\usepackage{amsthm}

\usepackage[capitalize,noabbrev]{cleveref}

\theoremstyle{plain}

\theoremstyle{definition}

\theoremstyle{remark}

\usepackage{wrapfig}
\usepackage{microtype}
\usepackage{graphicx}
\usepackage{booktabs} 
\usepackage{graphicx}
\usepackage{listings}
\usepackage{subcaption}
\usepackage{caption}
\usepackage{diagbox}
\usepackage{algorithm}
\usepackage{algpseudocode}
\usepackage{multirow}
\usepackage{enumitem}
\usepackage{colortbl} 
\usepackage{booktabs} 
\usepackage[skins]{tcolorbox} 
\usepackage{float}
\usepackage{authblk}
\setlist[itemize]{itemsep=2pt, topsep=0pt} 
\usepackage{stfloats}

\definecolor{main}{HTML}{5989cf}    
\definecolor{sub}{HTML}{cde4ff}     

\tcbset{
    sharp corners,
    colback = white,
    before skip = 0.2cm,    
    after skip = 0.5cm      
}                           

\newtcolorbox{boxB}[2][]{
    enhanced, 
    title=#1,
    fonttitle=\bfseries,
    coltitle=main,
    fontupper=\color{black}, 
    boxrule=1.5pt,
    colframe=main,
    rounded corners,
    arc=5pt, 
    colback=gray!20, 
    colbacktitle=white,
    attach boxed title to top left={yshift=-2mm, xshift=5mm},
    boxed title style={
        colframe=main,
        boxrule=1pt,
        rounded corners,
        arc=3pt,
        colback=white
    },
    #2
}
\newtcolorbox{boxA}{
    fontupper = \bf,
    boxrule = 1.5pt,
    colframe = black 
}

\title{CausalGraph2LLM: Evaluating LLMs for Causal Queries}

\author{%
  \textbf{Ivaxi Sheth$^{1}$, Bahare Fatemi$^{2}$, Mario Fritz$^{1}$} \\
 $^{1} $CISPA Helmholtz Center for Information Security, $^{2}$ Google Research\\
 \texttt{\{ivaxi.sheth,fritz\}@cispa.de, baharef@google.com} \\
 \\ \url{https://github.com/ivaxi0s/CausalGraph2LLM}
}

\begin{document}
\maketitle
\begin{abstract}
Causality is essential in scientific research, enabling researchers to interpret true relationships between variables. These causal relationships are often represented by causal graphs, which are directed acyclic graphs. With the recent advancements in Large Language Models (LLMs), there is an increasing interest in exploring their capabilities in causal reasoning and their potential use to hypothesize causal graphs. These tasks necessitate the LLMs to encode the causal graph effectively for subsequent downstream tasks. In this paper, we introduce \textbf{CausalGraph2LLM}, a comprehensive benchmark comprising over \textit{700k} queries across diverse causal graph settings to evaluate the causal reasoning capabilities of LLMs.  We categorize the causal queries into two types: graph-level and node-level queries. We benchmark both open-sourced and propriety models for our study. Our findings reveal that while LLMs show promise in this domain, they are highly sensitive to the encoding used. Even capable models like GPT-4 and Gemini-1.5 exhibit sensitivity to encoding, with deviations of about $60\%$. We further demonstrate this sensitivity for downstream causal intervention tasks. Moreover, we observe that LLMs can often display biases when presented with contextual information about a causal graph, potentially stemming from their parametric memory. 

\end{abstract}
\setcounter{tocdepth}{-1}

\section{Introduction}

The recent success of Large Language Models (LLMs)~\cite{brown2020language, achiam2023gpt, reid2024gemini} across various applications has opened up new avenues for their use beyond standard Natural Language Processing (NLP) tasks~\cite{srivastava2022beyond,wei2022chain}. Trained on massive corpora of structured and unstructured data~\cite{achiam2023gpt}, these models have demonstrated the ability to extract insights and exhibit emergent behaviors that can be harnessed for a wide range of applications~\cite{bubeck2023sparks, qi2023large,wang2023hypothesis, zhao2024expel}. 

Causal reasoning plays a pivotal role in guiding scientific research to establish causal relationships between different variables of an environment~\cite{pearl2009causality}. These relationships are often represented and modeled using causal graphs, which are directed and acyclic graphs. Traditionally, causal inference and discovery have been largely driven by observational data obtained through experiments~\cite{spirtes2016causal, nogueira2022methods, huang2020causal, cooper2013causal}. However, inferring causal graphs from observational data alone is a challenging problem~\cite{spirtes2016causal, brouillard2020differentiable}. This bottleneck of causal discovery has led to an increasing interest in the potential of LLMs to assist in this process~\cite{vashishtha2023causal, anonymous2023causal, liu2024discovery, ban2023query, ban2023causal, afonja2024llm4grn}. Therefore, the current paradigm for employing LLMs in causal discovery typically involves the use of metadata, particularly in the form of variable names to guide the models in identifying and interpreting causal relationships. Existing works utilize LLMs in various roles such as priors, critics, and post-processors for causality-related tasks. 

\begin{figure} [!t]
\centering
    \includegraphics[width=0.45\textwidth]{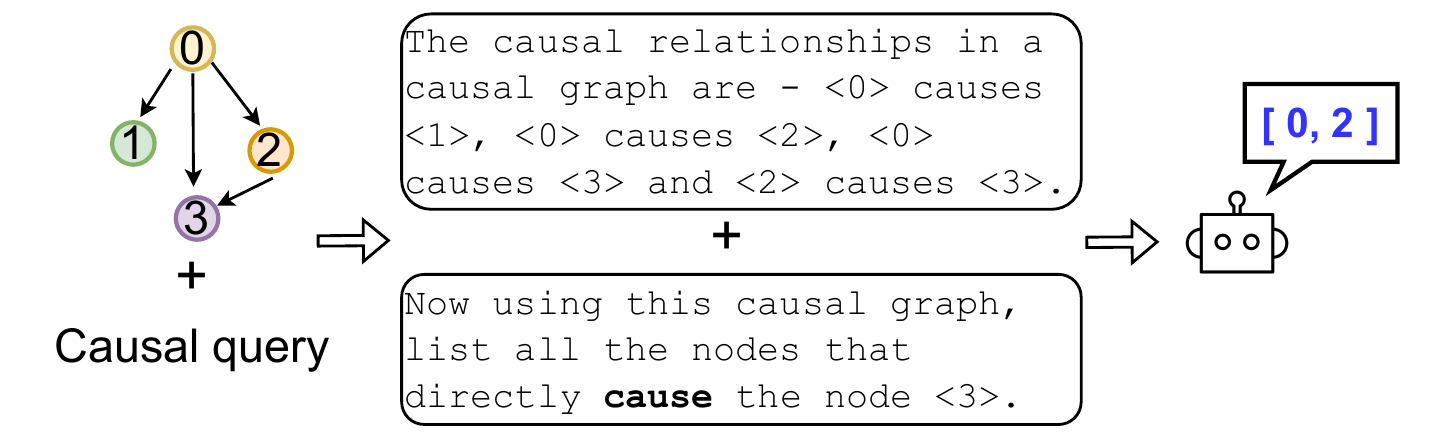}
          \caption{\textbf{CausalGraph2LLM:} Causal graphs are ingested into LLMs via prompt encoding strategies which are evaluated for causal queries.}
         \label{fig:teaser}
        \vspace{-2mm}
\end{figure}

While LLMs have demonstrated competitive performance~\cite{anonymous2023causal} against traditional data-driven methods, their effectiveness is constrained by their sequential text-based training paradigm. Current models typically necessitate a user to decompose their causal reasoning task into first textualizing a causal graph and then the task prompt. In essence, for downstream tasks, the LLMs need to be able to handle and manipulate textual representations of causal graphs effectively. This capability of LLMs being able to process causal graphs as text efficiently with any encoding is often assumed in current works. ~\citet{fatemi2023talk} have demonstrated sensitivity to prompts and encoding strategy for graphs, however, the focus of these works is on graph theory-based tasks, different from causal queries.
 
We challenge the assumption that LLMs can seamlessly encode causal graphs and evaluate their true capabilities in this area. By introducing the benchmark, we aim to shed light on the strengths and limitations of these models in encoding causal graphs. To fully harness the potential of LLMs for causal reasoning, it is crucial to understand not only the opportunities they present but also the risks and precautions necessary for their effective use. While LLMs can enhance our ability to discover and understand causal relationships, they may also propagate biases from their training data and their performance can vary based on the prompting strategy and task. Therefore, careful evaluation and consideration of potential biases and limitations are essential when using LLMs for causal reasoning. Considering the application of using LLMs as causal hypothesis generators~\cite{liu2024discovery, kiciman2023causal, ban2023causal, sheth2024hypothesizing}, it is imperative to evaluate their basic causal graph understanding capabilities before advancing to more complex tasks. Addressing any potential challenges early on can help in refining the models, making them more robust and effective as tools for causal reasoning and hypothesis generation. 

\paragraph{Contributions.} In this work, we aim to investigate the ability of LLMs to encode causal graphs and their effectiveness in assisting with causal reasoning tasks. We propose the benchmark, \emph{CausalGraph2LLM}, designed to evaluate LLMs on tasks related to understanding causal graphs. Our work is the first work to focus on the encoding strategies and sensitivities of LLMs in the context of causal graphs. We assess the performance of a variety of LLMs across a broad spectrum of tasks, each inspired by potential subtasks that LLMs might need to solve for a downstream task. This benchmark serves as a foundational reference for future works that use LLMs for causal reasoning-based tasks. 
Our contributions can be summarised as follows:

\begin{itemize}
    \setlength{\itemsep}{0pt} 
    \setlength{\parskip}{0pt} 
    \setlength{\parsep}{0pt}  

    \item We conduct a comprehensive study on various techniques to encode causal graphs into text for an LLM.
    \item We break down the task into several subtasks and graph-level and node-level queries to better understand the capabilities and limitations of LLMs in causal reasoning.
    \item Our work revealed biases in model performances when the contextual information was part of the pretraining data.
    \item We conduct extensive experiments on both open-source and closed models to uncover the limitations of LLMs in perfectly understanding causal graphs.
\end{itemize}

\section{Related Works}


Causal discovery and inference have predominantly been dominated by data-driven methods~\cite{spirtes2016causal}. However, due to the complexity of inferring causal structures, previous works have introduced priors on causal graphs in terms of interventions, domain expertise, edge existence, or ancestral constraints~\cite{constantinou2023impact, ban2023query, brouillard2020differentiable}. These priors help to reduce the search spaces of potential causal graphs. Recent advancements in LLMs have motivated the use of LLM-based priors and causal discovery~\cite{long2023can, cai2023knowledge, anonymous2023causal, jin2023cladder, kiciman2023causal}. Unlike data-driven methods, LLMs leverage causal variable names to evaluate the existence of edges between them, thereby constructing causal graphs. The rich parametric knowledge of LLMs has proven to be almost as effective in discovering causal structures as traditional data-driven methods~\cite{vashishtha2023causal,kiciman2023causal, afonja2024llm4grn}. These initial results have motivated the integration of LLMs as priors combined with different statistical causal discovery methods. For instance, \citet{vashishtha2023causal} used pairwise queries to discover the existence of edges between different causal variables and then applied methods such as PC~\cite{spirtes2001causation} to reorient the edges, whereas \citet{ban2023query} utilized LLM-based priors for scoring-based discovery methods. \citet{vashishtha2023causal} suggest triplet-based prompting strategies, and \citet{jiralerspong2024efficient} proposed reducing the prompting complexity by prompting in a depth-first search manner. More recently, \citet{anonymous2023causal} proposed an iterative collaboration between LLMs and structural causal models, where the LLM refines the output of SCMs. Another line of previous works~\cite{girju2002text, hassanzadeh2020causal,tan2023unicausal} explored the use of LLMs to discover potential causal structures from unstructured data.  ~\citet{chen2024clear} benchmarked the performance of different LLMs against various causal queries. 
Combined with external tools, ~\citet{jin2023can} demonstrated the use of LLMs for causal inference tasks, albeit on 3-4 node tasks. 
~\citet{wang2024causalbench} assesses causal graph understanding on causal tasks. In contrast, we focus on how encoding strategies and pretraining context impact reasoning performance on causal.

Most of these works assume a specific prompting strategy. However, it remains unclear which strategy would be most effective. In this paper, we aim to contribute to this line of research by benchmarking a variety of LLMs on a range of tasks related to causal graphs and exploring the effectiveness of different causal graph encoders.

\section{CausalGraph2LLM}
Understanding causal graphs is a crucial step in harnessing LLMs for tasks based on causal graphs. Our benchmark, CausalGraph2LLM, is designed to evaluate the proficiency of LLMs in interpreting and utilizing causal graphs, a skill that is vital for applications in causal inference and discovery. An overview of the benchmark is depicted in \autoref{fig:teaser}. By evaluating the ability of these models to process and comprehend the structure and implications of causal graphs, we aim to gain a deeper understanding of their potential and limitations in complex reasoning tasks.
\subsection{Preliminaries}
\begin{figure*} [!htb]
\centering
         \includegraphics[width=0.9\textwidth]{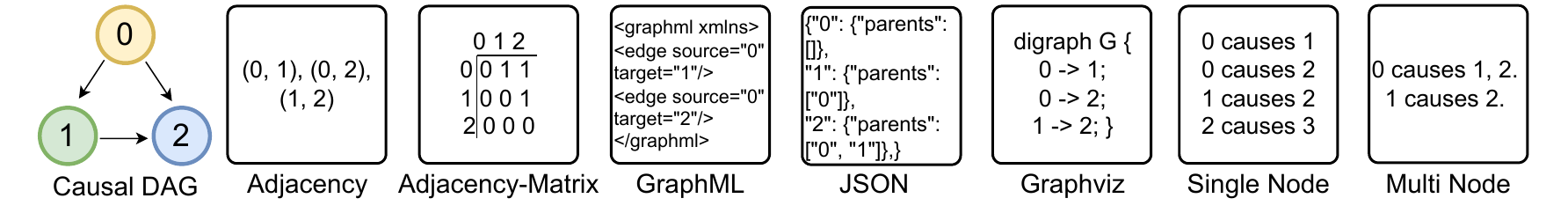}
          \caption{Different graph encoding functions for converting same causal graph to textual prompts, \(p: G \rightarrow P\). 
          }
         \label{fig:Prompts}
         \vspace{-2mm}
\end{figure*}

Causal graphs serve as an effective medium for conveying the perceived interactions among variables. These assumptions can be demonstrated in a directed acyclic graph (DAG), enabling researchers to infer which variables need to be controlled to reduce bias and identify those that could potentially introduce bias if controlled in the analysis. 
A causal graph is mathematically denoted as $G = (V, E)$, where $V$ is a finite set of vertices or nodes, represented as $\{v_1, v_2, \ldots, v_n\}$, with each $v_i$ signifying a distinct variable in the system. $E$ is a set of ordered pairs of vertices, denoted as $\{(v_i, v_j) \mid v_i, v_j \in V \text{ and } i \neq j\}$, with each pair $(v_i, v_j)$ representing a directed edge from node $v_i$ to node $v_j$. The presence of an edge from $v_i$ to $v_j$ signifies a causal effect of the variable represented by $v_i$ on the variable represented by $v_j$. Importantly, $G$ is a DAG, meaning that for any node $v_i$, there does not exist a directed sequence of edges that starts and ends at $v_i$.

\subsection{Prompting}

Instruction-tuned LLMs are gaining popularity for solving tasks using LLMs, driven by their ability to leverage pre-trained knowledge to infer causal structures by simply prompting these models. Therefore, we employ prompting in our approach. We benchmark the ability of LLMs to understand causal graphs by transforming the causal graph into a verbalized prompt. Given a causal graph G, we define a prompting function \(p: G \rightarrow P\), where \(P\) is the space of all possible prompts. This function transforms the graph into a verbalized format that the LLM can process. In this benchmark, we use various prompting transformations as used in current works or graph learning literature. We perform extensive experiments on 7 different encoding strategies. A brief overview of the encoding functions is illustrated in \autoref{fig:Prompts}. In Appendix \ref{app:prompting}, we go into the details for each encoding strategy. We hypothesize that popular graph representations can influence performance in downstream tasks. While the specific graph structures encountered during LLM pretraining are unknown, our benchmark aims to benchmark which graph representation enhances performance for LLMs.
The different encoders we use:
\begin{itemize}
    \setlength{\itemsep}{1pt} 
    \setlength{\parskip}{1pt} 
    \setlength{\parsep}{1pt}  

    \item \textbf{JSON} - This encoding represents the causal graph in a JSON format, capturing nodes and their causal relationships hierarchically~\cite{anonymous2023causal}.
    \item \textbf{Adjacency} - This encoding lists all edges in the graph, showing direct causal relationships between nodes~\cite{fatemi2023talk}.
    \item \textbf{Adjacency matrix} - This encoding uses a matrix to represent the graph, with each cell indicating the presence (1) or absence (0) of a direct causal relationship between nodes.
    \item \textbf{GraphML} - This encoding uses the GraphML format, a comprehensive XML-based format for graph representation.
    \item \textbf{Graphviz} - This encoding uses the DOT language, which can be visualized using Graphviz tools.
    \item \textbf{Single node} - This encoding lists each node with its direct effects in a straightforward textual description.
    \item \textbf{Multi node} - This encoding provides a multi-node description where each cause is followed by all its effects.
    \end{itemize}


\subsection{Tasks}
\begin{figure*} [!htb]
\centering
         \includegraphics[width=0.9\textwidth]{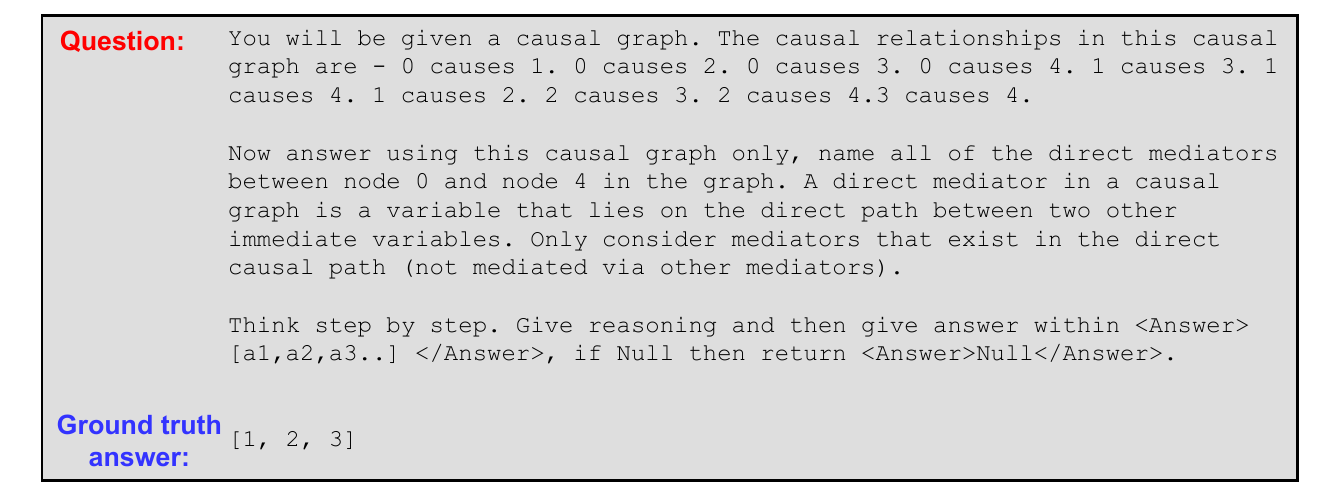}
          \caption{An example prompt with single node encoding, with mediator graph-level query.
          }
         \label{fig:Prompts}
         \vspace{-2mm}
\end{figure*}

For the benchmark, we consider various causality-based tasks that could serve as potential subtasks, each of which may be crucial for a language model's downstream performance in understanding causal graphs. Following the graph encoding prompt, a task question prompt is added. Each task is designed to evaluate an LLM's ability to interpret causal graphs from different aspects of causality. Given a causal graph $G = (V, E)$, we explore various types of relationships between nodes, each with its relevance to causality, causal inference, and causal discovery.
\paragraph{Child and Parent:} If there is a directed edge from node $v_i$ to node $v_j$ in $E$, then $v_i$ is the parent of $v_j$, and $v_j$ is the child of $v_i$. This relationship signifies a direct causal effect from the parent variable to the child variable. 
\paragraph{Source and Sink:} A source node is a node that has no incoming edges. It represents a variable that is not caused by any other variable in the system, often serving as the starting point of causal chains. A sink node is a node that has no outgoing edges. It represents a variable that does not cause any other variable in the system, often serving as the endpoint of causal chains. 
\paragraph{Mediator:} A mediator node in $V$ is a node that lies on the path between two other nodes in $E$. It represents a variable that mediates the causal effect from one variable to another, playing a key role in the propagation of causal effects.
\paragraph{Confounder:} A confounder node in $V$ is a node that has outgoing edges to two or more other nodes in $E$. It represents a variable that can induce spurious associations between its child variables if not properly controlled.

By evaluating the LLM's ability to identify and interpret these different types of relationships, we aim to gain a comprehensive understanding of its capabilities and limitations in the context of causal reasoning.

\subsubsection{Downstream Task}
We build upon the graph interventional effect task as proposed by ~\citet{kasetty2024evaluating} to evaluate the intervention reasoning abilities of an LLM as a downstream task. 

We expand the dataset to include larger graphs beyond 3 variables. In causal inference, interventions alter variable values within a causal graph, breaking their causal dependencies. This is represented using Pearl's do-calculus notation as \( \textit{do}(X = x) \), where \( X \) is the intervened variable and \( x \) is the set value. This intervention isolates \( X \) from its original causes, allowing for the analysis of the intervention's downstream effects. 

In a causal graph \( G = (V, E) \), where \( V \) is a set of variables and \( E \) is a set of directed edges representing causal relationships, applying \( \textit{do}(X = x) \) results in a modified graph where \( X \) is fixed at \( x \) and any incoming edges to \( X \) are removed. 

The task is to determine the intervention's impact on other variables. The LLM must infer whether the intervened variable $X$ causes changes in the other variables, based on the causal graph's structure. The LLM's task is to identify and quantify the intervention  $\textit{do}(X = x)$'s downstream effects on the variables $V\backslash{X}$.  This requires the LLM to interpret the causal graph and perform interventional reasoning. We evaluate this task by measuring the accuracy of LLM predictions.
\subsection{Experimental Setup}
\begin{figure*}[!ht]
\centering
     \begin{subfigure}[t]{0.32\textwidth}
         \centering
         \includegraphics[width=\textwidth]{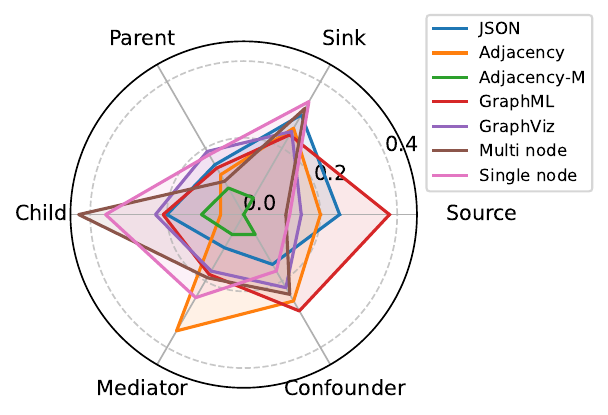}
         \caption{GritLM}
     \end{subfigure}
     \begin{subfigure}[t]{0.32\textwidth}
         \centering
         \includegraphics[width=\textwidth]{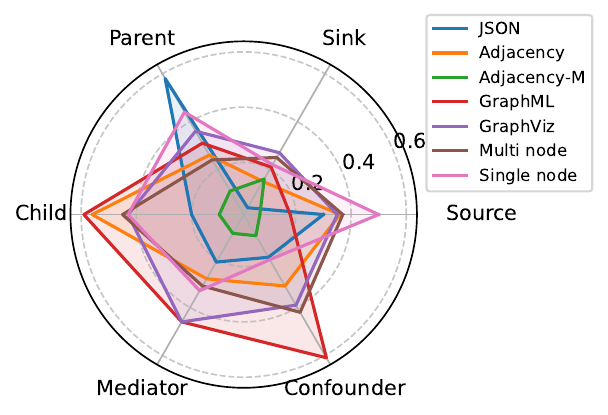}
         \caption{Mistral}
     \end{subfigure}
     \begin{subfigure}[t]{0.32\textwidth}
         \centering
         \includegraphics[width=\textwidth]{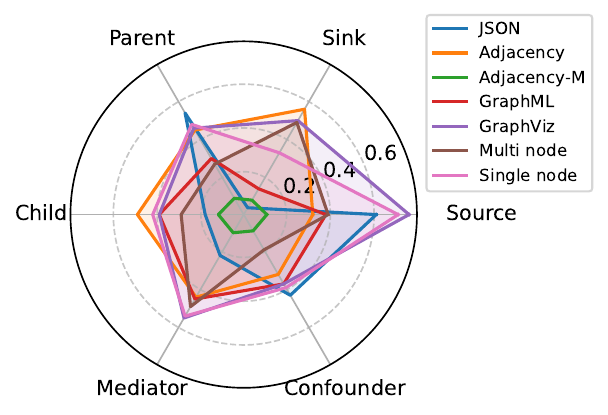}
         \caption{Mixtral}
     \end{subfigure}
     \begin{subfigure}[t]{0.32\textwidth}
         \centering
         \includegraphics[width=\textwidth]{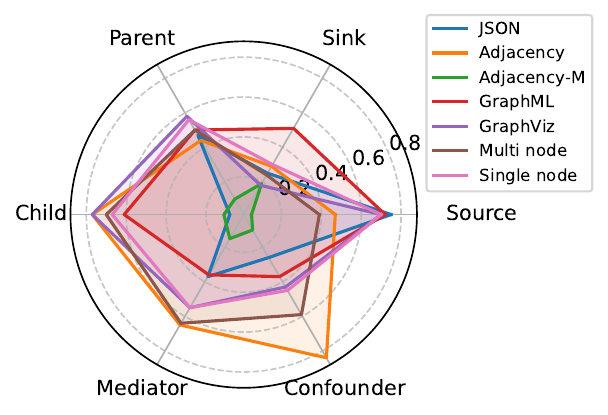}
         \caption{GPT-3.5}
     \end{subfigure}    
    \begin{subfigure}[t]{0.32\textwidth}
         \centering
         \includegraphics[width=\textwidth]{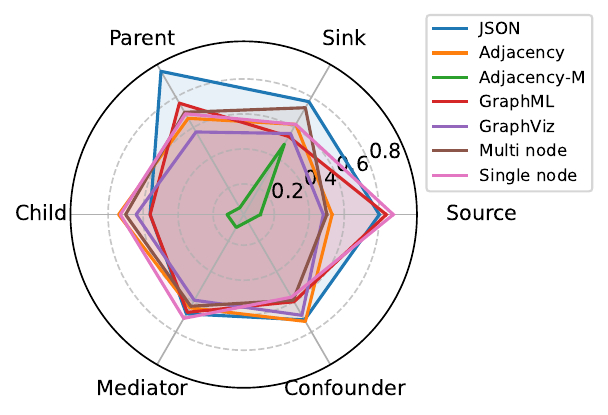}
         \caption{Gemini}
     \end{subfigure}     
     \begin{subfigure}[t]{0.32\textwidth}
         \centering
         \includegraphics[width=\textwidth]{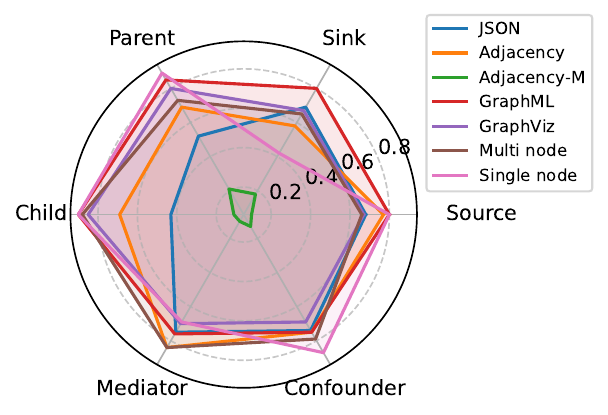}
         \caption{GPT-4}
     \end{subfigure}     

     \caption{Performance comparison across methods and encodings for graph-level queries.}
     \label{fig:spider_detail}
\end{figure*}

We evaluate our benchmark setup on a diverse array of graphs, including synthetic, and realistic contextual graphs. To control the complexity and structure of the graphs, we construct synthetic DAGs. Our benchmark also includes commonly used causal graphs from recent literature ~\cite{ban2023query, vashishtha2023causal, ban2023causal}, which incorporate contextual information i.e. variable semantics. We consider the causal graphs part of BNLearn repository - Insurance:${G}(27,52)$~\cite{binder1997adaptive}, and Alarm:${G}(37,46)$~\cite{beinlich1989alarm}. 

We evaluate the benchmark for various open-source and closed models. The models we use are  GritLM~\cite{muennighoff2024generative},
Mistral-7B-Instruct-v0.2~\cite{jiang2023mistral}, Mixtral-8x7BInstruct-v0.1~\cite{jiang2024mixtral}, Gemini~\cite{reid2024gemini}\footnote{The author affiliated with Google was responsible for the Gemini experiments and the authors affiliated with CISPA Helmholtz Center for Information Security were responsible for the rest experiments.}, GPT-3.5 ~\cite{brown2020language}, and GPT-4~\cite{achiam2023gpt}.

\section{Results}
This section presents our benchmark results on causal graph understanding through causal queries. We investigate how effectively LLMs can interpret and reason about causal graphs encoded in different formats, addressing both graph-level and node-level queries. Additionally, we explore biases introduced by graph contextual information. The variances are reported in Appendix \ref{app:variance} for brevity.

\subsection{Graph-level Queries}

To evaluate the baseline graph level causal queries, we prompt the LLMs with causal query tasks resembling those encountered in larger causal reasoning tasks with different encodings. We measure the performance of these queries using the F1 score.

\paragraph{LLMs may struggle with simple causal query tasks.}
From \autoref{fig:spider_detail}, we observe a range of performances across different models and encoding types, highlighting the variability in how well each LLM handles causal graph encoding and interpretation. Out of Source and Sink based queries, interestingly the LLM has stronger performance on performing source tasks. We ablate in Appendix \ref{app:order} and observe that the \textit{order} of causal graph description also has an impact on the performance of source and sink queries. This implies that the model's understanding of causal relationships may be influenced by the sequence in which information is presented. Tasks of greater complexity, such as identifying mediators, appear to be more challenging. This could be because the task of identifying a mediator intuitively involves breaking down the task into identifying 'child' and 'parent' elements, adding a layer of complexity to the task. We also conclude a correlation between graph complexity and performance in ~\autoref{appfig:nodes} and ~\autoref{appfig:edges}.
\paragraph{Average performance.} Observing the average performance for each model across different encodings suggests that the LLMs are highly sensitive to graph encoding. Adjacency-matrix encoding generally results in the lowest average performance across all models, despite being a popular format to represent causal graphs in code. 
\paragraph{High sensitivity to the causal graph representation.}
We observe that different encodings for the same causal graphs have different performances across each causal query. For instance, for the Mistral model, JSON encoding has the F1 score of $0.21$, however for GraphML or GraphViz encoding the performance increases to $0.46$ for the Mediator task. GPT-4 and Gemini 1.5 Pro perform exceptionally well with certain encodings like GraphML and JSON, respectively, indicating that these formats might align better with the potential pretraining of the model.  GritLM and Mistral show greater variability in their average performance, highlighting their sensitivity to the encoding methods used.  
\paragraph{Correlation between query and encodings.}
Some queries may seem easier due to the definition of the encoding and its potential alignment with the encoding. For instance, for JSON encoding, identifying parent nodes might be relatively easier for all LLMs. This could be because the JSON-based prompt used by ~\cite{anonymous2023causal} defines the dictionary by specifying the parents of each node. This alignment between the query and encoding likely facilitates the model's understanding of the causal relationships, resulting in improved performance on tasks involving parent nodes. This shows the importance of considering the encoding method coupled with the query when concerned with a causal graph-level reasoning task.

\subsection{Effect of pretraining knowledge}
In our previous experiment, we used synthetically generated causal graphs to assess the ability of LLMs to interpret and reason about causal relationships in a controlled setting. However, in this experiment, we aim to evaluate the impact of pretraining knowledge on the understanding of causal graphs. Current research often employs LLMs to extract causal priors by leveraging the semantic information embedded in variable names~\cite{anonymous2023causal, vashishtha2023causal, ban2023causal, sheth2024hypothesizing}. Their goal is to harness the knowledge and emergent reasoning abilities of LLMs to generate causal hypotheses. The primary motivation there is to harness the knowledge and emergent reasoning abilities of LLMs to generate causal hypotheses. Consequently, in this experiment, we specifically test causal graphs that incorporate contextual knowledge, allowing us to assess how pretraining influences LLMs' performance directly. We specifically consider two popular causal DAGs - Insurance~\cite{binder1997adaptive}, and Alarm~\cite{beinlich1989alarm}. These graphs were presented in two formats: one set featured contextual causal knowledge with semantically meaningful labels, and the other set consisted of the same graphs labeled with random identifiers. 

\begin{figure}
\centering
    \begin{subfigure}[t]{0.5\textwidth}
         \centering
         \includegraphics[width=0.8\textwidth]{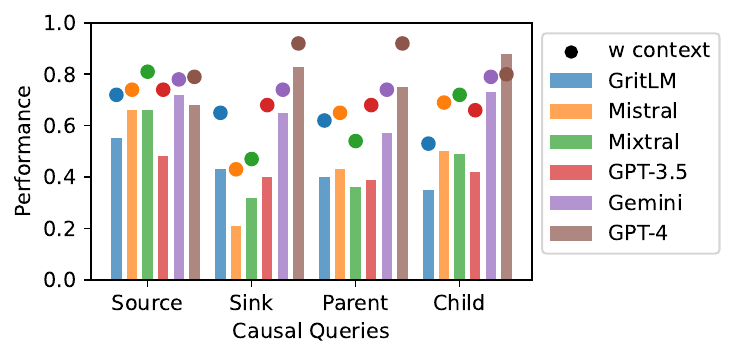}
         \vspace{-1mm}
         \caption{Insurance.}
     \end{subfigure}
     \begin{subfigure}[t]{0.5\textwidth}
         \centering
         \includegraphics[width=0.8\textwidth]{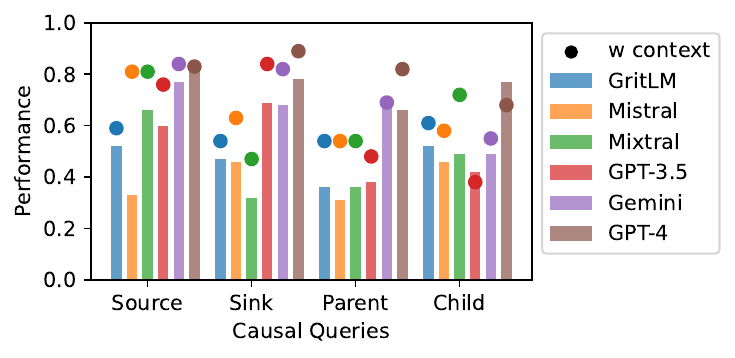} 
        \vspace{-1mm}
         \caption{Alarm.}
     \end{subfigure}
    \caption{Performance of different models across (a) Insurance and (b) Alarm graphs. Bars represent performance without context, while dots indicate performance with context. The results are averaged over different encodings for each model.}
    \label{fig:contextvsnocontext}
    \vspace{-2mm}
\end{figure}

From \autoref{fig:contextvsnocontext} we observe that giving contextual knowledge in terms of semantically meaningful causal variable names for the causal understanding tasks improves the performance across the models. This boost suggests that LLMs are effectively utilizing their pretraining on vast text corpora, where they have been exposed to a wide range of contexts and scenarios involving related variables. The semantics of the variable names likely help the models to enable a more accurate interpretation of the causal relationships by activating their parametric memory. 
\paragraph{Risks associated with reliance on contextual knowledge.}
While the improvement in performance with contextual knowledge is promising, it also raises some concerns. Primarily, the potential (over-) reliance on semantically meaningful variable names can introduce biases based on the language and cultural context inherent in the training data of the LLMs. We additionally observe that due to contextual knowledge, false positives were increased as in the case of GPT-4 for the Child query for the Insurance graph. This occurs because the LLM assumes there are more causal relationships than are specified by the causal graphs. This observation aligns with findings from~\cite{vashishtha2023causal}, where it was also observed that LLMs suggest more causal relationships as a causal prior in comparison to the ground truth causal graph. Moreover, when flipping the directions of the DAG to generate an anti-commonsense DAG, there is a drop in performance (see \autoref{tab:contextual}). This suggests that if a causal DAG with contextual information does not follow the pretraining of the LLM, due to inherited biases of the LLM, it may lead to more incorrect causal query inferences.

\subsection{Node-level queries are simpler for LLM}

In the above experiments, we considered \textit{graph-level} tasks only, that require the LLM to identify and list all instances of the mentioned node type present in a given causal graph. This requires the model to understand the entire graph structure and identify all nodes that follow the given task criteria (such as source, sink). In this experiment, we break down the graph-level tasks into binary queries for \textit{node-inspection} tasks to better understand the model's performance on simpler, more focused queries. 
\begin{figure}
    \centering
    \includegraphics[width=0.38 \textwidth]{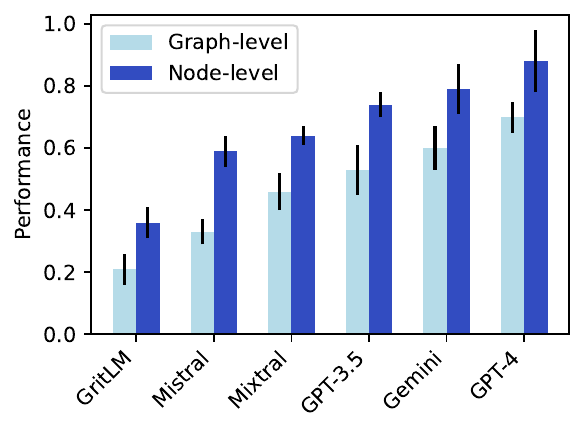}    
    \vspace{-2mm}
    \caption{Node-level vs. graph-level query performances.}
    \label{fig:binvs}
\end{figure}
The graph-level tasks are converted into binary queries reducing the processing load on the model. Each binary query asks whether a specific node in the graph is of a given type allowing LLM to focus on individual nodes. 

Our results, summarized in~\autoref{fig:binvs}, indicate that LLMs generally perform better on node-level query tasks. Lower performance for graph-level tasks can be attributed to the difficulty of the tasks. This could be explained as the graph-level queries require the LLM to maintain a holistic understanding of the graph's structure. Secondly, incorrectly identifying one node can lead to a cascade of errors, hence reducing LLM's ability to understand the entire graph structure. In contrast, \textit{node-inspection} tasks isolate each node-based query, reducing the impact of individual errors and leading to more accurate overall performance.
\subsubsection{Overestimation and Underestimation biases}

For the binary \textit{node-level} queries, we can further break down the results for each LLM and evaluate their false positives (FP) and false negatives (FN). This allows for an insight into the LLM to evaluate where the error truly comes from. False positives occur when the LLM incorrectly identifies a node as a specific type when it is not. False negatives occur when the LLM fails to identify a specific node type. We average the ratio of FP to FN across every task and embedding to report this ratio ($\tau$) for each model. A high $\tau$ ratio indicates that the model is more prone to false positives, meaning it frequently identifies nodes as a specific type even when they are not. Conversely, a low $\tau$ ratio (i.e., $\tau < 1$) would indicate a higher rate of false negatives, where the model fails to identify nodes that are 
of a specific type. This underestimation suggests that the model might be overly conservative in identifying causal relationships. 

\begin{figure}
    \centering
    \includegraphics[width=0.4 \textwidth]{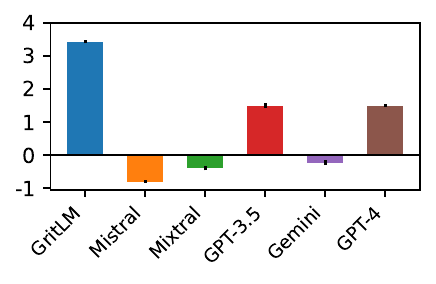}
    \caption{Evaluation of overestimation and underestimation biases. The models with positive values have high false positives and the models with negative values have high false negatives.}
    \label{fig:fnfp}
\end{figure}

From \autoref{fig:fnfp}, we observe that GritLM, GPT-3.5, and GPT-4 have $\tau > 1$ ratios. It suggests that the model tends to causal overestimation bias even without the influence of contextual information. Some of the recent works have also explored the overestimation phenomenon of LLMs~\cite{herrera-berg2023large, li2024your}. However, Gemini, Mistral, and Mixtral portray opposite behaviors, and their False Negative predictions are higher. Such biases could arise from their RLHF fine-tuning stage as well and require further investigation to explore such biases for causal queries.
\vspace{-6mm}
\subsection{Downstream task performance}
In this work, we aimed to benchmark LLM performance for various causal graph queries. From \autoref{tab:mainresult}, we observed that different encodings have varied effects across different causal tasks. In this experiment, however, we aim to observe this effect on a downstream task. 
\begin{figure}
    \centering
    \includegraphics[width=0.4 \textwidth]{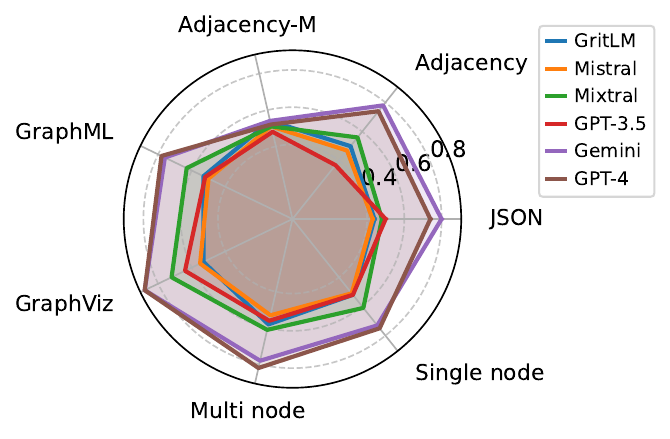}
    \caption{Downstream task performance.}
    \label{fig:downstream}
\end{figure}

From \autoref{fig:downstream}, we observed variability in model performances due to different encoding functions for downstream intervention tasks. At first glance, it may seem that adjacency-matrix encoding has higher performance in the intervention tasks as compared to graph-level queries. However, the random baseline is $0.50$, and the results here for the adjacency matrix are close to the random baseline even for the best-performing GPT-4 model.

\section{Discussion}

Our study, centered on zero-shot prompting, assesses how the comprehensive training of current models impacts their responses to causal queries. As shown in Table 1, models can handle causal queries reasonably well with appropriate encoding. However, performance varies based on the encoding used, likely due to differences in understanding the rich distribution within each encoding.
We hypothesize that fine-tuning could improve performance across different encodings by allowing models to adapt their pre-existing knowledge to specific causal queries. We tested this by fine-tuning the Mistral-7b model in \autoref{app:ft}. We observe the biggest increase in the performance of adjacency matrix encoding. 
In addition to textual encoding, we also explored visual graph encoding with more advanced models. In this approach, the LLM is prompted with an image of the causal graph. Interestingly, we found that the performance using visual encoding outperformed some of the textual encodings, although it did not emerge as the best approach (see \autoref{app:vision}). 

Despite the improvements brought about by fine-tuning, the performance indicates that there is still room for enhancement. Additionally, while fine-tuning proved effective, the influence of contextual knowledge and biases on performance remains an open question. We anticipate that this benchmark will serve as a valuable tool in addressing these questions and developing defence method against the biases.

\subsection{Key Takeaways}
Our findings highlight several critical aspects of causal graph understanding in LLMs, revealing both strengths and limitations.
\begin{enumerate}
    \item We observe up to a \textbf{60\% variation} in performance across different graph encodings. This suggests that model performance is not solely dependent on inherent reasoning abilities but also on how causal information is represented. Selecting the optimal encoding can significantly enhance LLM performance for specific causal tasks.

    \item Certain encodings, such as JSON for parent-child tasks, \textbf{align naturally with specific query} structures. This interaction between encoding format and query type has been largely overlooked in prior work. Adaptive encoding strategies  that optimize performance across a range of causal query types.

    \item Our results reveal a strong \textbf{pretraining-driven bias}: models perform significantly better when variable names align with patterns encountered in their training data. This finding raises concerns about applying LLMs to domains like medical causal graphs, where emerging research might deviate from pretraining knowledge.

    \item Tasks focusing on individual nodes, such as identifying mediators, are consistently easier than graph-level tasks.

    \item We observe \textbf{overestimation} (false positives) and \textbf{underestimation} (false negatives). This suggests that encoding choice not only affects overall performance but also introduces systematic errors, which could have significant implications for fields like policy modeling and epidemiology.

\end{enumerate}

\section{Conclusion}
With the increasing use of LLMs to assist with causal inference and discovery tasks, it is crucial to explore their potential and limitations. In this paper, we introduced CausalGraph2LLM, the first benchmark designed to evaluate how well LLMs encode and reason about causal DAGs across both graph-level and node-level queries. Our findings highlight not only the strengths of LLMs in handling causal queries but also the risks posed by pretraining biases, encoding sensitivity, and disparities in reasoning across different query types. These insights underscore the need for careful encoding choices, fine-tuning strategies, and bias mitigation techniques when applying LLMs to causal tasks. We hope this benchmark serves as a foundation for future research, driving improvements in LLM robustness, interpretability, and applicability for causal DAG analysis.

\section{Limitations and Future Work}
The scope of the evaluation is primarily limited to synthetic and well-known causal graphs, which may not fully capture the complexity of real-world causal graphs. We presented 6 diverse tasks, which can be built upon for future work. Future work can expand the diversity of causal graphs and models evaluated, develop more robust encoding techniques, and explore methods to mitigate contextual biases. Enhancing the models' ability to handle complex tasks and improving downstream task performance will also be crucial. Additionally, a deeper investigation into biased sources can provide a more nuanced understanding of LLM capabilities in causal inference. Given the modular nature of the benchmark, we aim to continue to build up this benchmark to assess newer models as they come.

\section{Ethics and Risks}

All of the datasets used are publicly available. Our implementation utilizes the PyTorch 1.12 framework, an open-source library. Our research is conducted per the licensing agreements of the Mistral-7B, GPT-3.5, and GPT-4 models. We ran our experiments on A100 Nvidia GPU and via OpenAI API. 

While this benchmark can serve as a research tool for studying LLMs’ capabilities and improving their robustness, it should not be interpreted as an endorsement of LLMs as reliable causal inference tools. Users should apply caution and rigorous validation when leveraging LLMs for causal analysis, ensuring that their outputs are interpreted critically rather than taken at face value. 
\section{Acknowledgements}
This work was partially funded by ELSA – European Lighthouse on Secure and Safe AI funded by the European Union under grant agreement No. 101070617. Moreover, the computation resources used in this work are supported by the Helmholtz
Association’s Initiative and Networking Fund on the HAICORE@FZJ partition. Views and opinions expressed are however those of the authors only and do not necessarily reflect those of the European Union or European Commission. Neither the European Union nor the European Commission can be held responsible for them. 

\newpage

\bibliography{custom}

\setcounter{tocdepth}{2}
\appendix
\onecolumn
\section{Reproduciblility}

We will release our code, prompts, evaluation setup, and all models’ outputs of our experiments. For reproducibility, we used temperature $0$ and top-p value as $1$ across all of the models. We also mentioned the snapshot of the model used. We have also included the prompts and examples below. Our code will be made public post the anonymity period.

The Alarm and Insurance datasets are under CC BY-SA 3.0 which allows us to modify the datasets for benchmarking freely. Our benchmark will be released under the CC BY-SA License. 

Mistral and GritLM were run on 1 A100 GPU whereas Mixtral was run on 8 A100 GPUs. Since we used off-the shelf LLM, each graph-level experiment took no more than 30 minutes to run (longer for mediator, child, parent, confounder whereas source and sink took $\approx$ 3 mins to run). All of the experiments for each model took $\approx 38$ hours. GPT-3.5 and GPT-4 were accessed via API. 
\subsection{Dataset descriptions}
The datasets used can be divided into two: 1. Contextual i.e realistic datasets and 2. synthetic datasets.

We use the two real-world-based datasets. These are semi-synthetic datasets available from the BNLearn library. The first graph, named \textbf{Alarm}, is a well-known benchmark in the field of causal inference. The Alarm dataset (see Figure 11) is designed to model the relationships and dependencies in an intensive care unit (ICU) monitoring system. It includes variables such as heart rate, blood pressure, and other vital signs, making it a complex and realistic representation of medical data. This dataset is particularly useful for evaluating the ability of LLMs to handle intricate causal relationships in a medical high-stakes environment.

The second dataset, \textbf{Insurance}, is another widely used benchmark that models the risk factors and dependencies in the insurance domain. This graph (see Figure 12) includes variables related to policyholders, such as age, driving history, and vehicle type, and their relationships to insurance claims and premiums. The Insurance dataset provides a different context from the medical domain, allowing us to assess the versatility of LLMs in understanding and reasoning about causal relationships in a financial setting.

\subsection{Synthetic dataset}
In addition to real-world-based datasets, we created synthetic datasets with varying levels of difficulty to rigorously evaluate the performance of LLMs. These synthetic datasets were designed to systematically vary in complexity by adjusting the number of nodes and edges in the causal graphs. This variation allows us to assess how well the models handle different levels of graph complexity and density. The synthetic datasets serve as a controlled environment to test the models' ability to interpret and reason about causal relationships under varying conditions. By incrementally increasing the number of edges while keeping the number of nodes constant, we can observe how the models' performance scales with the complexity of the causal structure. This approach provides valuable insights into the strengths and limitations of LLMs in handling more intricate causal graphs, which is crucial for understanding their potential applications in real-world scenarios. For the experiments, we synthesized graphs with 20 and 30 nodes. For each of these node variables, we experimented with different densities of nodes. Hence we had density = 1 x nodes, 1.5 x nodes and 2 x nodes. 

\subsection{Dataset statistics}

\textbf{Scale of the benchmark.} The benchmark covers a wide range of graph sizes and tasks, including both graph-level and node-level queries, ensuring robust evaluation. Our benchmark includes a total of 70,638 + 36,184 = 106,822 (contextual + synthetic) queries for each encoding. Hence the total number of queries across all of the encoding is 747,754.

\noindent\textbf{Utility.} Our benchmark evaluates both synthetic and contextual graphs, allowing users to assess models' performance in general and domain-specific scenarios.

Below is a summary of the total number of queries for each task:

\begin{table}[H]
\tiny
\centering
\begin{tabular}{c|l|cc|cc|cc|cc|cc|cc}
\toprule
\textbf{Graph Type} & \textbf{Query Type} & \multicolumn{2}{c|}{\textbf{Source}} & \multicolumn{2}{c|}{\textbf{Sink}} & \multicolumn{2}{c|}{\textbf{Mediator}} & \multicolumn{2}{c|}{\textbf{Confounder}} & \multicolumn{2}{c|}{\textbf{Child}} & \multicolumn{2}{c}{\textbf{Parent}}  \\
 & & Graph & Node & Graph & Node & Graph & Node & Graph & Node & Graph & Node & Graph & Node \\
\midrule
\multirow{2}{*}{\textbf{Contextual}} & Graph-Level & 1 & 64 & 1 & 64 & 1071 & 32085 & 1071 & 32085 & 64 & 2034 & 64 & 2034 \\
 & Node-Level & 64 & 64 & 64 & 64 & 32085 & 32085 & 32085 & 32085 & 2034 & 2034 & 2034 & 2034 \\
\midrule
\multirow{2}{*}{\textbf{Synthetic}} & Graph-Level & 2 & 60 & 2 & 60 & 670 & 15960 & 670 & 15960 & 60 & 1340 & 60 & 1340 \\
 & Node-Level & 60 & 60 & 60 & 60 & 15960 & 15960 & 15960 & 15960 & 1340 & 1340 & 1340 & 1340 \\
\bottomrule
\end{tabular}
\caption{Comparison of Query Types in Contextual and Synthetic Graphs}
\label{tab:graph_comparison}
\end{table}

\section{Prompting Strategies}
\label{app:prompting}

We query the llms by prompting. below we go through different types of prompts. the prompt can be broken into:
\begin{enumerate}
    \item Causal query explanation
    \item Encoding
    \item Query
\end{enumerate}
\subsection{Causal query explanation}
\begin{boxB}[Source]{}
A source node in a causal graph is a variable that does not have any incoming edges, meaning it is not caused by any other variable in the graph.
\end{boxB}

\begin{boxB}[Sink]{}
A sink node in a causal graph is a variable that does not have any children in the graph, meaning it is not caused by any other variables in the system.
\end{boxB}

\begin{boxB}[Direct Mediator]{}
A direct mediator in a causal graph is a variable that lies on the direct path between two other immediate variables. Only consider mediators that exist in the direct causal path (not mediated via other mediators).
\end{boxB}

\begin{boxB}[Confounder]{}
A confounder in a causal graph is a variable that influences both the cause and the effect variables. It is a common cause for both the dependent and independent variables.
\end{boxB}

\begin{boxB}[Parents]{}
What nodes are the direct causes of Node X?
\end{boxB}

\begin{boxB}[Child]{}
What nodes are directly caused by Node X?
\end{boxB}
\subsection{Encoding}
\begin{boxB}[Adjacency]{}
( 0 ,  1 ) \\
( 0 ,  2 ) \\
( 1 ,  3 ) \\
( 2 ,  3 ) \\
( 2 ,  4 ) \\
( 3 ,  4 ) \\
( 0 ,  3 ) \\
( 1 ,  4 ) \\
( 0 ,  4 ) \\
( 1 ,  2 )
\end{boxB}
\begin{boxB}[Adjacency Matrix]{}
\[
\begin{array}{c|ccccc}
  & 0 & 1 & 2 & 3 & 4 \\
\hline
0 & 0 & 1 & 1 & 1 & 1 \\
1 & 0 & 0 & 1 & 1 & 1 \\
2 & 0 & 0 & 0 & 1 & 1 \\
3 & 0 & 0 & 0 & 0 & 1 \\
4 & 0 & 0 & 0 & 0 & 0 \\
\end{array}
\]
\end{boxB}
\begin{boxB}[GraphML]{}
\begin{verbatim}
<?xml version="1.0" encoding="UTF-8"?>
<graphml xmlns="http://graphml.graphdrawing.org/xmlns">
  <graph edgedefault="directed">
    <node id="0"/>
    <node id="1"/>
    <node id="2"/>
    <node id="3"/>
    <node id="4"/>
    <edge source="0" target="1"/>
    <edge source="0" target="2"/>
    <edge source="1" target="3"/>
    <edge source="2" target="3"/>
    <edge source="2" target="4"/>
    <edge source="3" target="4"/>
    <edge source="0" target="3"/>
    <edge source="1" target="4"/>
    <edge source="0" target="4"/>
    <edge source="1" target="2"/>
  </graph>
</graphml>
\end{verbatim}
\end{boxB}
\begin{boxB}[GraphViz]{}
\begin{verbatim}
digraph G {
  0 -> 1;
  0 -> 2;
  1 -> 3;
  2 -> 3;
  2 -> 4;
  3 -> 4;
  0 -> 3;
  1 -> 4;
  0 -> 4;
  1 -> 2;
}
\end{verbatim}
\end{boxB}
\begin{boxB}[JSON]{}
\begin{verbatim}
{
    "0": {
        "parents": []
    },
    "1": {
        "parents": [
            "0"
        ]
    },
    "2": {
        "parents": [
            "0",
            "1"
        ]
    },
    "3": {
        "parents": [
            "0",
            "1",
            "2"
        ]
    },
    "4": {
        "parents": [
            "0",
            "1",
            "2",
            "3"
        ]
    }
}
\end{verbatim}

\end{boxB}
\begin{boxB}[Multi node]{}

0 causes 1, 2, 3, 4.
1 causes 3, 4, 2.
2 causes 3, 4.
3 causes 4.
\end{boxB}
\begin{boxB}[Single node]{}

0 causes 1. 
0 causes 2.
0 causes 3.
0 causes 4.
1 causes 3.
1 causes 4.
1 causes 2.
2 causes 3.
2 causes 4.
3 causes 4.
\end{boxB}

\subsection{Prompt template}
For further prompt templates, please check the codebase. 
\begin{boxB}[Graph-level query prompt]{}
Hello. You will be given a causal graph. The causal relationships in this causal graph are - [causal-graph-based-encoding]. Now answer using this causal graph only, name all of the [node-type] in the graph. [node-type-description]. Think step by step. Give reasoning and then give answer within <Answer> [a1,a2,a3..] </Answer>, if Null then return <Answer>Null</Answer>.
\end{boxB}

\begin{boxB}[Node-level query prompt]{}
Hello. You will be given a causal graph. The causal relationships in this causal graph are - [causal-graph-encodingbased]. Now answer using this causal graph only, is [nodeX] a [node-type] in the graph. [node-type-description]. Think step by step. Give reasoning and then give answer within <Answer> Yes/No </Answer>.
\end{boxB}

\begin{boxB}[\textit{Example:} Single node: graph-level: Mediator]{}
Hello. You will be given a causal graph. The causal relationships in this causal graph are - 0 causes 1. 
0 causes 2.
0 causes 3.
0 causes 4.
1 causes 3.
1 causes 4.
1 causes 2.
2 causes 3.
2 causes 4.
3 causes 4. Now answer using this causal graph only, name all of the direct mediators between node 0 and node 4 in the graph. A direct mediator in a causal graph is a variable that lies on the direct path between two other immediate variables. Only consider mediators that exist in the direct causal path (not mediated via other mediators). 
Think step by step. Give reasoning and then give answer within <Answer> [a1,a2,a3..] </Answer>, if Null then return <Answer>Null</Answer>.
\end{boxB}

\section{Experiments}
\subsection{Variance}
\label{app:variance}
\begin{table}[htb!]
\small
\centering
\begin{tabular}{c|l|*{6}{p{0.9cm}|}{p{1.2cm}}}
\toprule
\textbf{Model} & Enc&\textbf{{Source}} & \textbf{{Sink}} & \textbf{{Parent}} & \textbf{{Child}} & \textbf{\tiny{Mediator}} & \textbf{\tiny{Confounder}} & \textit{{Avg}} \\
\toprule
\multirow{7}{*}{\rotatebox[origin=c]{90}{\textbf{GritLM}}} & JSON & $\underset{\pm0.07}{0.25}$ & $\underset{\pm0.05}{0.30}$ & $\underset{\pm0.02}{0.15}$ & $\underset{\pm0.07}{0.20}$ & $\underset{\pm0.08}{0.10}$ & $\underset{\pm0.07}{0.15}$ & 0.19\tiny{$\pm$0.10} \\
&Adjacency & $\underset{\pm0.03}{0.20}$ & $\underset{\pm0.04}{0.26}$ & $\underset{\pm0.01}{0.12}$ & $\underset{\pm0.02}{0.06}$ & $\underset{\pm0.05}{0.35}$ & $\underset{\pm0.04}{0.26}$ & 0.20\tiny{$\pm$0.12} \\
&Adjacency-M & $\underset{\pm0.00}{0.00}$ & $\underset{\pm0.01}{0.05}$ & $\underset{\pm0.01}{0.08}$ & $\underset{\pm0.02}{0.11}$ & $\underset{\pm0.01}{0.06}$ & $\underset{\pm0.01}{0.06}$ & 0.06\tiny{$\pm$0.03} \\
&GraphML & $\underset{\pm0.06}{0.38}$ & $\underset{\pm0.04}{0.24}$ & $\underset{\pm0.03}{0.14}$ & $\underset{\pm0.05}{0.21}$ & $\underset{\pm0.04}{0.18}$ & $\underset{\pm0.05}{0.29}$ & 0.24\tiny{$\pm$0.08} \\
&GraphViz & $\underset{\pm0.03}{0.15}$ & $\underset{\pm0.05}{0.25}$ & $\underset{\pm0.04}{0.19}$ & $\underset{\pm0.04}{0.23}$ & $\underset{\pm0.03}{0.17}$ & $\underset{\pm0.04}{0.22}$ & 0.20\tiny{$\pm$0.03} \\
&Multi node & $\underset{\pm0.02}{0.11}$ & $\underset{\pm0.06}{0.32}$ & $\underset{\pm0.02}{0.10}$ & $\underset{\pm0.08}{0.43}$ & $\underset{\pm0.04}{0.19}$ & $\underset{\pm0.05}{0.24}$ & 0.23\tiny{$\pm$0.12} \\
&Single node & $\underset{\pm0.03}{0.12}$ & $\underset{\pm0.06}{0.34}$ & $\underset{\pm0.04}{0.18}$ & $\underset{\pm0.07}{0.36}$ & $\underset{\pm0.05}{0.25}$ & $\underset{\pm0.04}{0.17}$ & 0.23\tiny{$\pm$0.10} \\
 
 \midrule
\multirow{7}{*}{\rotatebox[origin=c]{90}{\textbf{Mistral}}} & JSON & $\underset{\pm0.03}{0.30}$ & $\underset{\pm0.01}{0.04}$ & $\underset{\pm0.06}{0.58}$ & $\underset{\pm0.02}{0.20}$ & $\underset{\pm0.02}{0.21}$ & $\underset{\pm0.02}{0.19}$ & 0.25\tiny{$\pm$0.18} \\
&Adjacency & $\underset{\pm0.04}{0.36}$ & $\underset{\pm0.02}{0.15}$ & $\underset{\pm0.03}{0.26}$ & $\underset{\pm0.06}{0.56}$ & $\underset{\pm0.03}{0.28}$ & $\underset{\pm0.03}{0.31}$ & 0.32\tiny{$\pm$0.13} \\
&Adjacency-M & $\underset{\pm0.01}{0.07}$ & $\underset{\pm0.02}{0.16}$ & $\underset{\pm0.01}{0.11}$ & $\underset{\pm0.01}{0.10}$ & $\underset{\pm0.01}{0.09}$ & $\underset{\pm0.01}{0.10}$ & 0.10\tiny{$\pm$0.03} \\
&GraphML & $\underset{\pm0.02}{0.18}$ & $\underset{\pm0.02}{0.21}$ & $\underset{\pm0.03}{0.31}$ & $\underset{\pm0.06}{0.59}$ & $\underset{\pm0.05}{0.46}$ & $\underset{\pm0.06}{0.61}$ & \underline{0.39\tiny{$\pm$0.18}} \\
&GraphViz & $\underset{\pm0.04}{0.35}$ & $\underset{\pm0.03}{0.27}$ & $\underset{\pm0.04}{0.36}$ & $\underset{\pm0.04}{0.43}$ & $\underset{\pm0.05}{0.46}$ & $\underset{\pm0.04}{0.39}$ & 0.37\tiny{$\pm$0.06} \\
&Multi node & $\underset{\pm0.04}{0.37}$ & $\underset{\pm0.03}{0.25}$ & $\underset{\pm0.02}{0.24}$ & $\underset{\pm0.05}{0.45}$ & $\underset{\pm0.03}{0.31}$ & $\underset{\pm0.04}{0.42}$ & 0.34\tiny{$\pm$0.08} \\
&Single node & $\underset{\pm0.05}{0.50}$ & $\underset{\pm0.02}{0.22}$ & $\underset{\pm0.04}{0.44}$ & $\underset{\pm0.04}{0.43}$ & $\underset{\pm0.03}{0.33}$ & $\underset{\pm0.02}{0.20}$ & 0.35\tiny{$\pm$0.12} \\

\midrule
\multirow{7}{*}{\rotatebox[origin=c]{90}{\textbf{Mixtral}}} & JSON & $\underset{\pm0.06}{0.61}$ & $\underset{\pm0.01}{0.04}$ & $\underset{\pm0.05}{0.54}$ & $\underset{\pm0.02}{0.18}$ & $\underset{\pm0.02}{0.22}$ & $\underset{\pm0.04}{0.43}$ & 0.33\tiny{$\pm$0.22} \\
&Adjacency & $\underset{\pm0.03}{0.32}$ & $\underset{\pm0.05}{0.56}$ & $\underset{\pm0.04}{0.45}$ & $\underset{\pm0.05}{0.49}$ & $\underset{\pm0.04}{0.44}$ & $\underset{\pm0.03}{0.32}$ & 0.43\tiny{$\pm$0.09} \\
&Adjacency-M & $\underset{\pm0.01}{0.11}$ & $\underset{\pm0.01}{0.08}$ & $\underset{\pm0.01}{0.09}$ & $\underset{\pm0.01}{0.12}$ & $\underset{\pm0.01}{0.10}$ & $\underset{\pm0.01}{0.09}$ & 0.10\tiny{$\pm$0.01} \\
&GraphML & $\underset{\pm0.04}{0.38}$ & $\underset{\pm0.01}{0.14}$ & $\underset{\pm0.03}{0.30}$ & $\underset{\pm0.04}{0.39}$ & $\underset{\pm0.04}{0.45}$ & $\underset{\pm0.04}{0.37}$ & 0.34\tiny{$\pm$0.10} \\
&GraphViz & $\underset{\pm0.07}{0.76}$ & $\underset{\pm0.05}{0.50}$ & $\underset{\pm0.04}{0.46}$ & $\underset{\pm0.04}{0.39}$ & $\underset{\pm0.05}{0.55}$ & $\underset{\pm0.04}{0.37}$ & \underline{0.50\tiny{$\pm$0.14}} \\
&Multi node & $\underset{\pm0.04}{0.39}$ & $\underset{\pm0.05}{0.49}$ & $\underset{\pm0.03}{0.27}$ & $\underset{\pm0.03}{0.29}$ & $\underset{\pm0.05}{0.49}$ & $\underset{\pm0.02}{0.19}$ & 0.35\tiny{$\pm$0.12} \\
&Single node & $\underset{\pm0.07}{0.71}$ & $\underset{\pm0.03}{0.33}$ & $\underset{\pm0.05}{0.48}$ & $\underset{\pm0.04}{0.42}$ & $\underset{\pm0.05}{0.54}$ & $\underset{\pm0.04}{0.39}$ & 0.48\tiny{$\pm$0.13} \\

\midrule
\multirow{7}{*}{\rotatebox[origin=c]{90}{\textbf{GPT-3.5}}} & JSON & $\underset{\pm0.07}{0.75}$ & $\underset{\pm0.03}{0.25}$ & $\underset{\pm0.05}{0.47}$ & $\underset{\pm0.01}{0.08}$ & $\underset{\pm0.04}{0.37}$ & $\underset{\pm0.03}{0.26}$ & 0.36\tiny{$\pm$0.23} \\
&Adjacency & $\underset{\pm0.05}{0.47}$ & $\underset{\pm0.03}{0.29}$ & $\underset{\pm0.04}{0.44}$ & $\underset{\pm0.08}{0.77}$ & $\underset{\pm0.07}{0.65}$ & $\underset{\pm0.09}{0.84}$ & \underline{0.57\tiny{$\pm$0.21}} \\
&Adjacency-M & $\underset{\pm0.01}{0.05}$ & $\underset{\pm0.02}{0.19}$ & $\underset{\pm0.01}{0.10}$ & $\underset{\pm0.01}{0.11}$ & $\underset{\pm0.02}{0.15}$ & $\underset{\pm0.01}{0.10}$ & 0.12\tiny{$\pm$0.11} \\
&GraphML & $\underset{\pm0.07}{0.72}$ & $\underset{\pm0.05}{0.51}$ & $\underset{\pm0.05}{0.50}$ & $\underset{\pm0.06}{0.61}$ & $\underset{\pm0.04}{0.36}$ & $\underset{\pm0.04}{0.37}$ & 0.51\tiny{$\pm$0.13} \\
&GraphViz & $\underset{\pm0.07}{0.70}$ & $\underset{\pm0.02}{0.18}$ & $\underset{\pm0.06}{0.58}$ & $\underset{\pm0.08}{0.77}$ & $\underset{\pm0.06}{0.55}$ & $\underset{\pm0.04}{0.43}$ & 0.53\tiny{$\pm$0.12} \\
&Multi node & $\underset{\pm0.04}{0.39}$ & $\underset{\pm0.02}{0.24}$ & $\underset{\pm0.05}{0.50}$ & $\underset{\pm0.07}{0.70}$ & $\underset{\pm0.06}{0.64}$ & $\underset{\pm0.06}{0.59}$ & 0.51\tiny{$\pm$0.17} \\
&Single node & $\underset{\pm0.07}{0.70}$ & $\underset{\pm0.03}{0.30}$ & $\underset{\pm0.06}{0.56}$ & $\underset{\pm0.07}{0.67}$ & $\underset{\pm0.06}{0.55}$ & $\underset{\pm0.05}{0.45}$ & 0.54\tiny{$\pm$0.14} \\
 
 \midrule

\multirow{7}{*}{\rotatebox[origin=c]{90}{\textbf{Gemini}}}&JSON & $\underset{\pm0.08}{0.80}$ & $\underset{\pm0.08}{0.77}$ & $\underset{\pm0.10}{0.97}$ & $\underset{\pm0.06}{0.56}$ & $\underset{\pm0.07}{0.68}$ & $\underset{\pm0.07}{0.72}$ & \underline{0.76\tiny{$\pm$0.13}} \\
&Adjacency & $\underset{\pm0.05}{0.53}$ & $\underset{\pm0.06}{0.62}$ & $\underset{\pm0.07}{0.66}$ & $\underset{\pm0.07}{0.74}$ & $\underset{\pm0.06}{0.64}$ & $\underset{\pm0.07}{0.73}$ & 0.66\tiny{$\pm$0.07} \\
&Adjacency-M & $\underset{\pm0.01}{0.12}$ & $\underset{\pm0.05}{0.49}$ & $\underset{\pm0.01}{0.07}$ & $\underset{\pm0.01}{0.12}$ & $\underset{\pm0.01}{0.11}$ & $\underset{\pm0.01}{0.07}$ & 0.22\tiny{$\pm$0.16} \\
&GraphML & $\underset{\pm0.08}{0.84}$ & $\underset{\pm0.05}{0.54}$ & $\underset{\pm0.08}{0.76}$ & $\underset{\pm0.06}{0.56}$ & $\underset{\pm0.07}{0.67}$ & $\underset{\pm0.06}{0.60}$ & 0.67\tiny{$\pm$0.11} \\
&GraphViz & $\underset{\pm0.05}{0.48}$ & $\underset{\pm0.06}{0.56}$ & $\underset{\pm0.06}{0.57}$ & $\underset{\pm0.06}{0.64}$ & $\underset{\pm0.06}{0.59}$ & $\underset{\pm0.07}{0.69}$ & 0.58\tiny{$\pm$0.07} \\
&Multi node & $\underset{\pm0.05}{0.50}$ & $\underset{\pm0.07}{0.73}$ & $\underset{\pm0.07}{0.70}$ & $\underset{\pm0.07}{0.70}$ & $\underset{\pm0.06}{0.63}$ & $\underset{\pm0.06}{0.59}$ & 0.64\tiny{$\pm$0.08} \\
&Single node & $\underset{\pm0.09}{0.88}$ & $\underset{\pm0.06}{0.62}$ & $\underset{\pm0.07}{0.69}$ & $\underset{\pm0.07}{0.73}$ & $\underset{\pm0.07}{0.71}$ & $\underset{\pm0.06}{0.57}$ & 0.71\tiny{$\pm$0.10}\\
\midrule

\multirow{7}{*}{\rotatebox[origin=c]{90}{\textbf{GPT-4}}} &JSON & $\underset{\pm0.07}{0.68}$ & $\underset{\pm0.07}{0.69}$ & $\underset{\pm0.05}{0.52}$ & $\underset{\pm0.04}{0.43}$ & $\underset{\pm0.08}{0.75}$ & $\underset{\pm0.07}{0.74}$ & 0.80\tiny{$\pm$0.13} \\
&Adjacency & $\underset{\pm0.08}{0.77}$ & $\underset{\pm0.06}{0.58}$ & $\underset{\pm0.07}{0.69}$ & $\underset{\pm0.07}{0.69}$ & $\underset{\pm0.08}{0.84}$ & $\underset{\pm0.08}{0.75}$ & 0.73\tiny{$\pm$0.09} \\
&Adjacency-M & $\underset{\pm0.01}{0.10}$ & $\underset{\pm0.02}{0.18}$ & $\underset{\pm0.02}{0.21}$ & $\underset{\pm0.01}{0.11}$ & $\underset{\pm0.01}{0.10}$ & $\underset{\pm0.01}{0.13}$ & 0.14\tiny{$\pm$0.04} \\
&GraphML & $\underset{\pm0.08}{0.80}$ & $\underset{\pm0.08}{0.80}$ & $\underset{\pm0.09}{0.85}$ & $\underset{\pm0.09}{0.90}$ & $\underset{\pm0.08}{0.76}$ & $\underset{\pm0.08}{0.75}$ & \underline{0.81\tiny{$\pm$0.05}} \\
&GraphViz & $\underset{\pm0.07}{0.67}$ & $\underset{\pm0.07}{0.67}$ & $\underset{\pm0.08}{0.80}$ & $\underset{\pm0.09}{0.85}$ & $\underset{\pm0.07}{0.70}$ & $\underset{\pm0.07}{0.69}$ & 0.71\tiny{$\pm$0.07} \\
&Multi node & $\underset{\pm0.07}{0.66}$ & $\underset{\pm0.07}{0.65}$ & $\underset{\pm0.07}{0.73}$ & $\underset{\pm0.09}{0.88}$ & $\underset{\pm0.08}{0.84}$ & $\underset{\pm0.08}{0.79}$ & 0.75\tiny{$\pm$0.09} \\
&Single node & $\underset{\pm0.08}{0.80}$ & $\underset{\pm0.04}{0.42}$ & $\underset{\pm0.09}{0.89}$ & $\underset{\pm0.09}{0.90}$ & $\underset{\pm0.07}{0.69}$ & $\underset{\pm0.09}{0.87}$ & 0.77\tiny{$\pm$0.18} \\

\midrule
\end{tabular}
\caption{Performance comparison across methods and encodings. }
\label{app:tab:mainresult}
\vspace{-4mm}
\end{table}

\begin{table*}[H]
\small
\centering
\begin{tabular}{l|*{7}{p{1.16cm}|}}
\toprule
& {\tiny{JSON}} & {\tiny{Adjacency}} & {\tiny{Adjacency-M}} & {\tiny{GraphML}} & {\tiny{GraphViz}}& {\tiny{Multi node}} & {\tiny{Single node}} \\
\midrule
GritLM & $\underset{\pm0.04}{0.44}$ & $\underset{\pm0.05}{0.50}$ & $\underset{\pm0.05}{0.54}$ & $\underset{\pm0.05}{0.53}$ & $\underset{\pm0.05}{0.53}$ & $\underset{\pm0.06}{0.58}$ & $\underset{\pm0.05}{0.54}$ \\
Mistral & $\underset{\pm0.04}{0.43}$ & $\underset{\pm0.05}{0.47}$ & $\underset{\pm0.05}{0.51}$ & $\underset{\pm0.05}{0.50}$ & $\underset{\pm0.05}{0.55}$ & $\underset{\pm0.05}{0.53}$ & $\underset{\pm0.06}{0.58}$ \\
Mixtral & $\underset{\pm0.05}{0.48}$ & $\underset{\pm0.06}{0.56}$ & $\underset{\pm0.05}{0.51}$ & $\underset{\pm0.06}{0.63}$ & $\underset{\pm0.07}{0.72}$ & $\underset{\pm0.06}{0.61}$ & $\underset{\pm0.06}{0.58}$ \\
GPT-3.5 & $\underset{\pm0.05}{0.50}$ & $\underset{\pm0.04}{0.37}$ & $\underset{\pm0.05}{0.48}$ & $\underset{\pm0.05}{0.52}$ & $\underset{\pm0.06}{0.64}$ & $\underset{\pm0.06}{0.56}$ & $\underset{\pm0.06}{0.58}$ \\
Gemini & $\underset{\pm0.08}{0.80}$ & $\underset{\pm0.08}{0.78}$ & $\underset{\pm0.05}{0.54}$ & $\underset{\pm0.07}{0.76}$ & $\underset{\pm0.04}{0.88}$ & $\underset{\pm0.08}{0.78}$ & $\underset{\pm0.06}{0.63}$ \\
GPT-4 & $\underset{\pm0.07}{0.74}$ & $\underset{\pm0.07}{0.74}$ & $\underset{\pm0.05}{0.52}$ & $\underset{\pm0.08}{0.78}$ & $\underset{\pm0.03}{0.88}$ & $\underset{\pm0.04}{0.82}$ & $\underset{\pm0.08}{0.77}$ \\
\midrule
\end{tabular}
\caption{Sensitivity to encoding for downstream intervention analysis.}
\label{tab:int}
\vspace{-5mm}
\end{table*}

\begin{figure}[h]
\centering
         \includegraphics[width=0.4\textwidth]{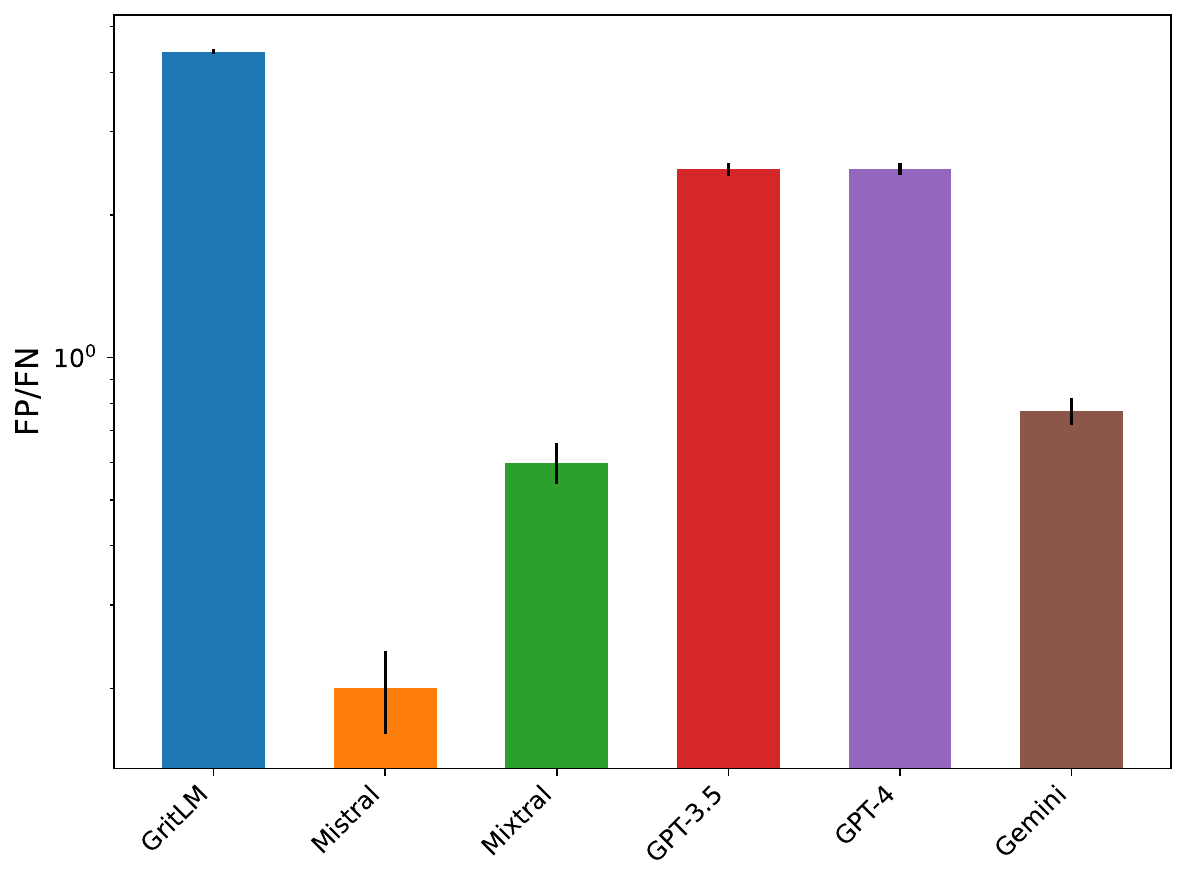}
          \caption{Evaluation of over- and underestimation biases. }
         \label{fig:ins}
\end{figure}

\subsection{Ordering of nodes matter for causal queries}
\label{app:order}
In BFS, the traversal starts from the source nodes, while in BFS-R, the traversal begins from the sink nodes. The values in the table represent the performance of the models on the tasks, with higher values indicating better performance.

The results show that the traversal order significantly impacts the performance of the models. For instance, GritLM performs better on source tasks when the traversal is in BFS order, while it performs better on sink tasks when the traversal is in BFS-R order. This pattern is consistent across all models, suggesting that BFS is more suitable for identifying source nodes, while BFS-R is more suitable for identifying sink nodes.
\begin{table}[H]
\small
\centering
\begin{tabular}{c|l|cc|cc}
\toprule
\textbf{D} & \textbf{Model} & \multicolumn{2}{c|}{\textbf{Source}} & \multicolumn{2}{c}{\textbf{Sink}}  \\
 & & BFS & BFS-R  & BFS  & BFS-R \\
 \midrule
\multirow{7}{*}{\rotatebox[origin=c]{90}{\textbf{Synthetic}}}  
& GritLM & 0.18 & 0.24 & 0.27 & 0.0.47 \\
& Mistral & 0.32 & 0.26 & 0.21 & 0.39 \\
& Mixtral & 0.48 & 0.40 & 0.31 & 0.44 \\
& GPT-3.5 & 0.57 & 0.48 & 0.31 & 0.64 \\
& Gemini & 0.65 & 0.54 & 0.62 & 0.82 \\
& GPT-4 & 0.68 & 0.57 & 0.61 & 0.89 \\
\midrule
\end{tabular}
\caption{Comparing the order for prompts, BFS means it starts from source and BFS-R means it starts from sinks. }
\label{tab:order}
\vspace{-4mm}
\end{table}

\subsection{Downstream performance under/over estimation bias}
\begin{figure}[h]
    \centering
         \includegraphics[width=0.4\textwidth]{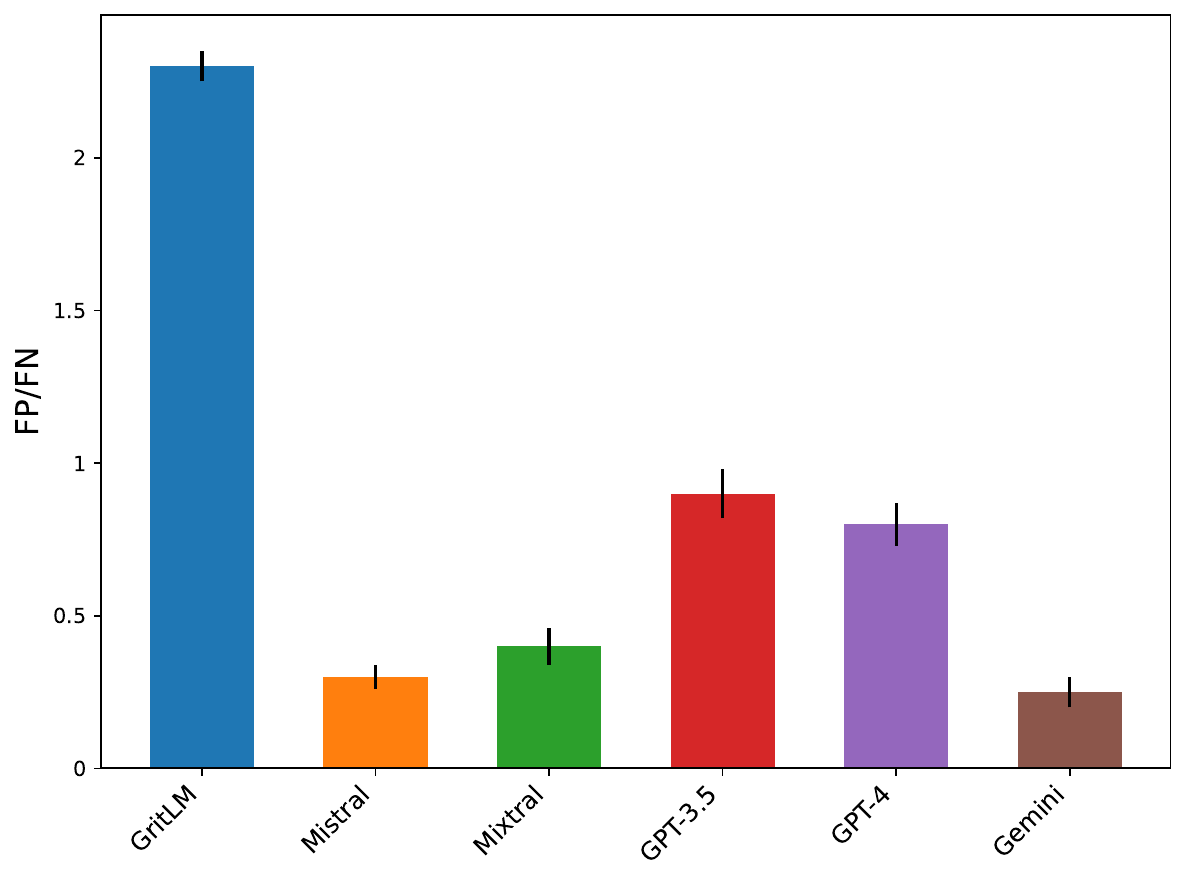}
          \caption{Evaluation of over- and underestimation biases for downstream task. }
         \label{appfig:bias}
\end{figure}

In the main paper, we analyzed over and underestimation bias for the binary node inspection task. We can conduct a similar analysis on the downstream task. Here, we observe a similar trend to the estimation biases in Section 4.3.1. Notably, GPT-3.5 and GPT-4 usually have FP/FN ratios closer to 1. See Figure~\ref{appfig:bias}. 

\subsection{Effect of node explanations}

In our experimental setup, we took an approach to defining each task for every metric. This was primarily due to the varying terminologies used in causal inference across different academic circles. For instance, what some researchers might refer to as a 'source', others might call a 'root'. To avoid any potential confusion, we provided clear definitions for each term used in our causal queries. 

Since pretraining for each model was not known, this adds an element of uncertainty to the task. To counteract this, we explicitly mentioned the query in our experiments. We conducted a set of preliminary experiments without an explanation of the query to demonstrate its effectiveness. The results showed a decrease in model performance, suggesting that providing explicit direction in the form of a mentioned query can be beneficial. 

\begin{table}[H]
\centering
\begin{tabular}{c|l|*{6}{p{1.2cm}|}}
\toprule
\textbf{Model} & \textbf{Enc}&\textbf{{Source}} & \textbf{{Sink}} & \textbf{{Parent}} & \textbf{{Child}} & \textbf{{Mediator}} & \textbf{{Confounder}} \\
\toprule
\multirow{7}{*}{\rotatebox[origin=c]{90}{\textbf{GPT-3.5}}} & JSON & 0.52 & 0.25 & 0.47 & 0.08 & 0.30 & 0.31 \\
 & Adjacency & 0.32 & 0.26 & 0.44 & 0.65 & 0.72 & 0.51 \\
 & Adjacency-M & 0.06 & 0.15 & 0.10 & 0.11 & 0.08 & 0.12 \\
 & GraphML & 0.34 & 0.38 & 0.50 & 0.61 & 0.37 & 0.39 \\
 & GraphViz & 0.42 & 0.19 & 0.58 & 0.77 & 0.52 & 0.28 \\
 & Multi node & 0.39 & 0.24 & 0.50 & 0.70 & 0.64 & 0.27 \\
 & Single node & 0.45 & 0.27 & 0.56 & 0.67 & 0.39 & 0.50 \\ \midrule
 \end{tabular}
\caption{Performance comparison across methods and encodings for GPT-3.5 without causal query explanations.}
\label{app:tab:nodeexplanation}
\end{table}

\subsection{Node complexity}
Graph complexity can significantly impact the performance of LLMs in processing and understanding structured data. One critical factor contributing to this complexity is the number of nodes within a graph. As the number of nodes increases, the structural complexity of the graph grows, introducing additional dependencies and relationships that the model must learn. Additionally, a denser graph with more edges introduces more pathways and connections, making it more challenging for the model to infer accurate relationships.

\begin{figure}[H]
    \centering
         \includegraphics[width=0.6\textwidth]{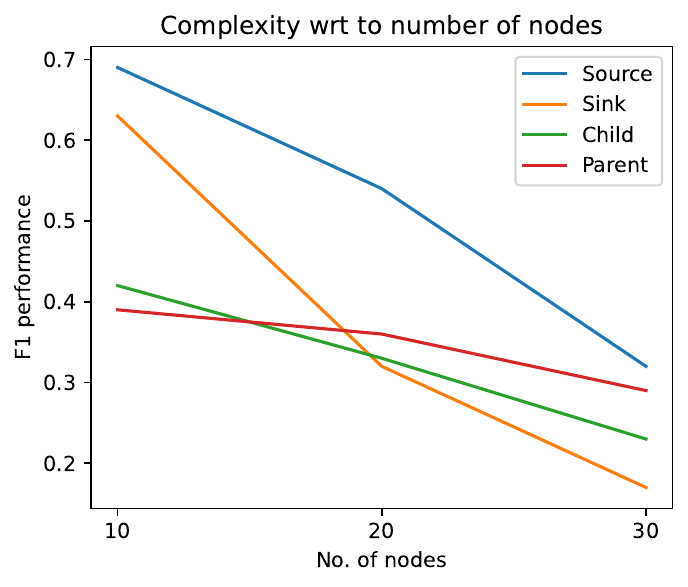}
          \caption{With an increase in graph complexity by increasing the number of nodes, we observe poorer performance of the LLM - Mistral model. }
         \label{appfig:nodes}
\end{figure}

\begin{figure}[H]
    \centering
         \includegraphics[width=0.6\textwidth]{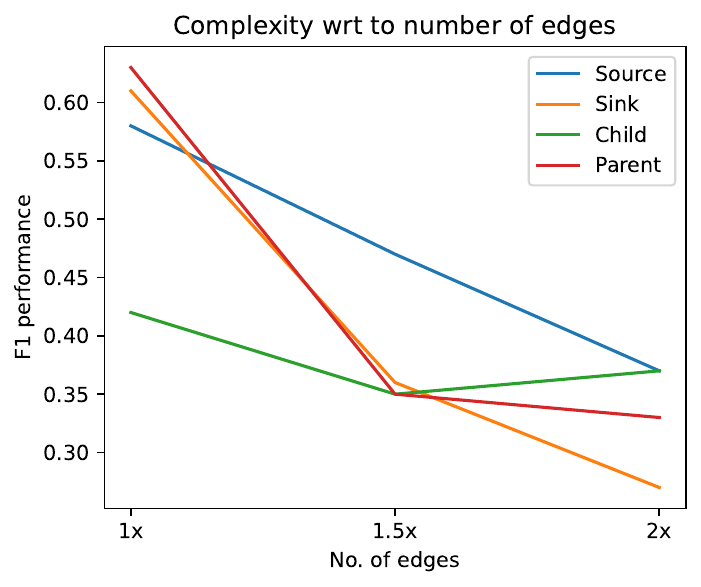}
          \caption{With an increase in graph complexity by increasing the number of edges, we observe poorer performance of the LLM - Mistral model. }
         \label{appfig:edges}
\end{figure}




\subsection{Multimodal models}
In this work we focus on textual encodings into LLMs, however with the developments of multimodal models, we can test LLM's ability to answer causal queries when presented with image inputs. We performed our experiment on GPT-4 model with T=0. Future works can be built upon to test better image inputs for multimodal models. 

\begin{table}[H]
    \centering
    \begin{tabular}{llllll}
    \toprule
        Source & Sink & Child & Parent & Mediator & Confounder \\ \midrule
        0.58 & 0.62 & 0.71 & 0.65 & 0.58 & 0.63 \\ \bottomrule
    \end{tabular}
    \caption{Performance metrics for vision GPT-4.}
    \label{app:vision}
\end{table}

\subsection{Effect of finetuning}
In this paper, we focused on zero-shot prompting as the current models have billions of trainable parameters and have been trained on a plethora of training data, potentially including causal graphs. We hence aimed to evaluate how this reflects in the causal queries. Additionally, most current methods utilize LLMs without fine-tuning for causal discovery queries, and our study aimed to replicate this environment to provide a realistic benchmark. We performed QLORA on Mistral 7b specifically on synthetic datasets. As expected, we observed an increase in the performance with finetuning.

\begin{table}[H]
    \centering
    \begin{tabular}{l|l|l|l|l|l|l}
        \toprule
        ~ & Source & Sink & Parent & Child & Mediator & Confounder  \\ \midrule
        JSON & 0.30 & 0.04 & 0.58 & 0.20 & 0.21 & 0.19  \\ 
        JSON FT & 0.63 & 0.36 & 0.73 & 0.42 & 0.33 & 0.44   \\  \midrule
         Adjacency & 0.36 & 0.15 & 0.26 & 0.56 & 0.28 &  0.31\\
        Adjacency FT & 0.53 & 0.44 & 0.71 & 0.46 & 0.41 &  0.52 \\ \midrule
        Adjacency-M & 0.07 & 0.16 & 0.11 &0.10  &0.09  & 0.10 \\
        Adjacency-M FT & 0.47 & 0.31 & 0.30 &0.38  &0.33  & 0.37 \\ \midrule
        GraphML & 0.18 & 0.21 & 0.31 & 0.59 & 0.46 & 0.61  \\ 
        GraphML FT & 0.47 & 0.42 & 0.55 & 0.73 & 0.68 & 0.73  \\ \midrule 
        Multi node & 0.37 & 0.25 & 0.24 & 0.45 & 0.31 & 0.42\\
        Multi node FT & 0.66 & 0.68 & 0.48 & 0.71 & 0.62 & 0.53 \\ \midrule

        Single node & 0.50 & 0.22 & 0.44 & 0.43 & 0.33 & 0.20 \\
        Single node FT  & 0.81 & 0.58 & 0.69 & 0.75 & 0.53 & 0.49 \\ \bottomrule

    \end{tabular}
    \caption{Effect of finetuning Mistral 7b model for different graph encoding.}
    \label{app:ft}
\end{table}

\subsection{Anti-commonsense context results}
In this experiment, we essentially flipped the arrows of the DAG. We observed that the LLM showed a decrease in performance showing its reliance on background parametric knowledge for reasoning tasks.

\begin{table}[H]
\small
\centering
\begin{tabular}{c|l|cc|cc|cc|cc}
\toprule
\textbf{Enc} & \textbf{Model} & \multicolumn{2}{c|}{\textbf{Source}} & \multicolumn{2}{c|}{\textbf{Sink}} & \multicolumn{2}{c|}{\textbf{Parent}} & \multicolumn{2}{c}{\textbf{Child}}  \\
 & & w/o  & anti  & w/o  & anti  & w/o  & anti  & w/o  & anti  \\
\midrule
\multirow{7}{*}{\rotatebox[origin=c]{90}{\textbf{Insurance}}}  
& GritLM & 0.55 & 0.38 & 0.43 & 0.21 & 0.40 & 0.18 & 0.35 & 0.17 \\

& Mistral & 0.66 & 0.58 & 0.21 & -0.01 & 0.43 & 0.21 & 0.50 & 0.31 \\

& Mixtral & 0.66 & 0.51 & 0.32 & 0.17 & 0.36 & 0.18 & 0.49 & 0.26 \\

& GPT-3.5 & 0.48 & 0.22 & 0.40 & 0.12 & 0.39 & 0.10 & 0.42 & 0.18 \\

& Gemini & 0.72 & 0.66 & 0.65 & 0.56 & 0.57 & 0.40 & 0.73 & 0.67 \\

& GPT-4 & 0.68 & 0.57 & 0.83 & 0.74 & 0.75 & 0.58 & 0.88 & 0.96  \\

\midrule

\multirow{7}{*}{\rotatebox[origin=c]{90}{\textbf{Alarm}}} 
& GritLM & 0.52 & 0.45 & 0.47 & 0.40 & 0.36 & 0.18 & 0.52 & 0.43 \\

& Mistral & 0.33 & 0.25 & 0.46 & 0.29 & 0.31 & 0.28 & 0.46 & 0.34 \\

& Mixtral & 0.66 & 0.51 & 0.32 & 0.17 & 0.36 & 0.18 & 0.49 & 0.26 \\

& GPT-3.5 & 0.60 & 0.44 & 0.69 & 0.54 & 0.38 & 0.28 & 0.42 & 0.46 \\

& Gemini & 0.77 & 0.70 & 0.68 & 0.54 & 0.71 & 0.73 & 0.49 & 0.43 \\

& GPT-4 & 0.82 & 0.81 & 0.78 & 0.67 & 0.66 & 0.50 & 0.77 & 0.86 \\

\midrule
\end{tabular}
\caption{Performance of different models across Alarm and Insurance graphs. w/o - without context anti - with reversed contextual variables. The results are averages across the encodings.}
\label{tab:contextual}
\vspace{-4mm}
\end{table}

\section{Contextual Graphs}
\begin{figure*}[htb!]
\centering
         \includegraphics[width=1.0\textwidth]{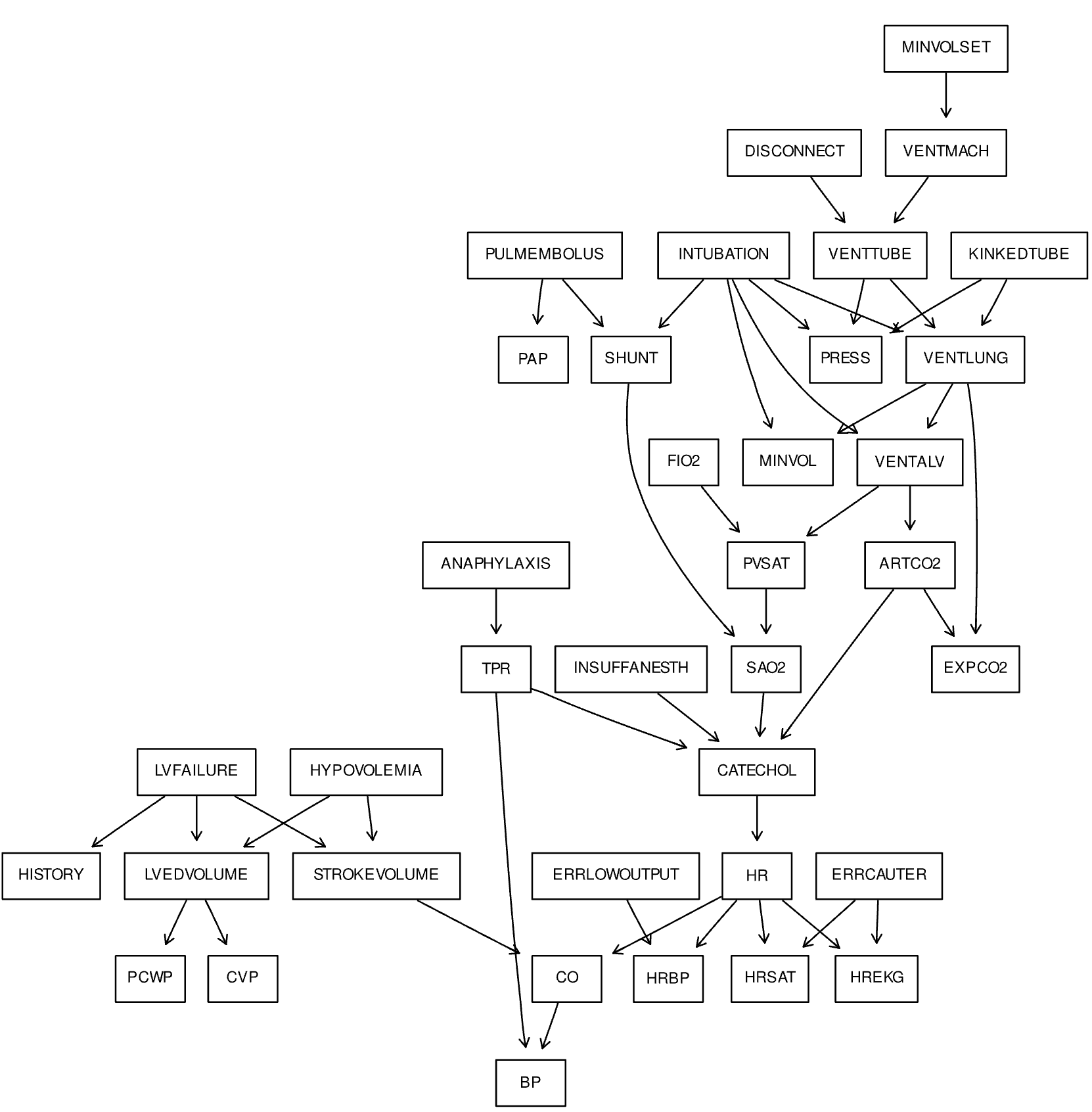}
          \caption{Alarm causal graph }
         \label{fig:ala}
\end{figure*}

\begin{figure*}[htb!]
\centering
         \includegraphics[width=1.0\textwidth]{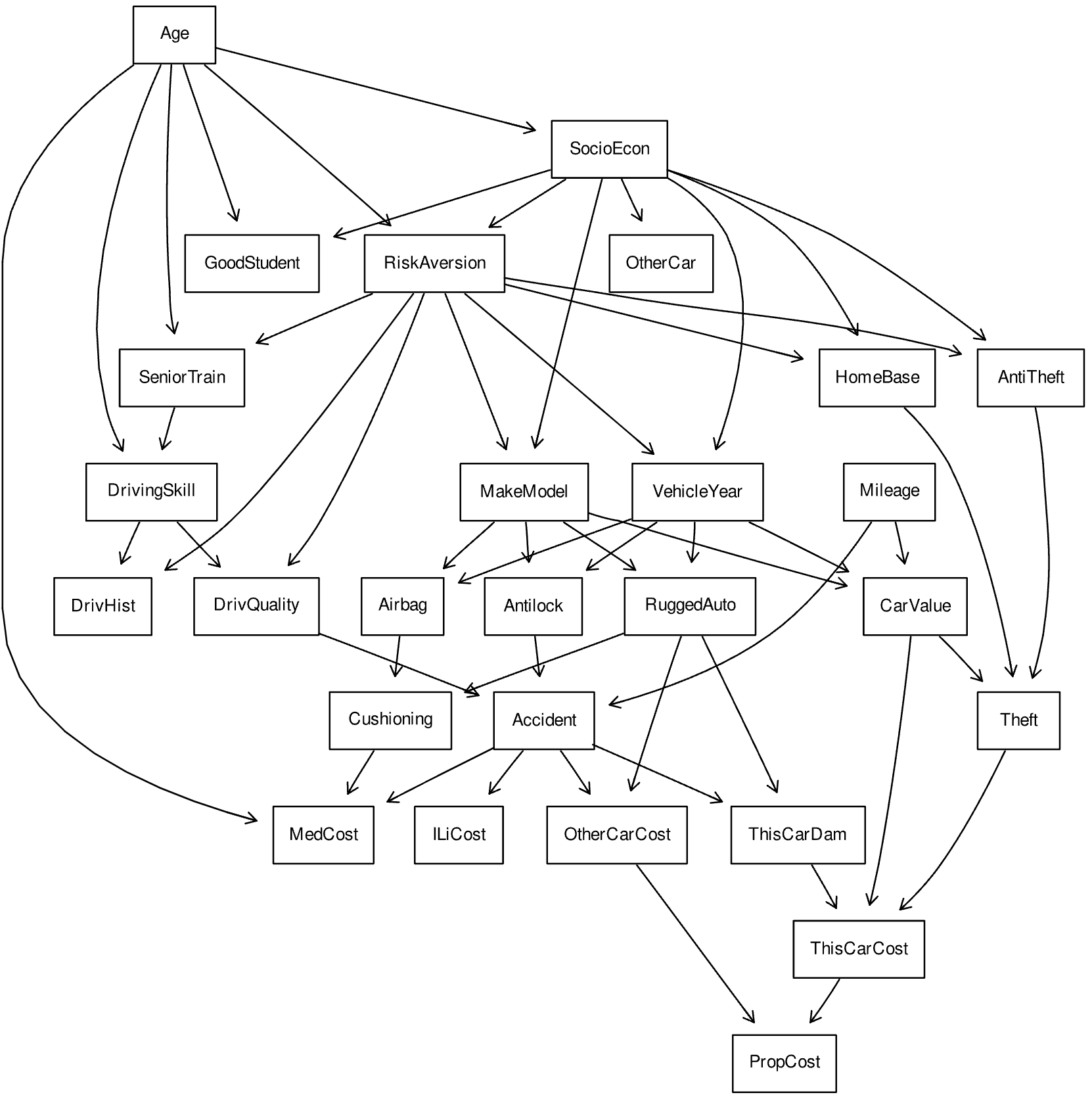}
          \caption{Insurance causal graph }
         \label{fig:ins}
\end{figure*}

\

\end{document}